%% file: main.tex
\documentclass{article}
\usepackage[nonatbib,final]{neurips_2022}
\usepackage{natbib}
\usepackage[hidelinks]{hyperref} 
\usepackage[utf8]{inputenc} 
\usepackage[T1]{fontenc}   
\usepackage{amsfonts}
\usepackage{mathtools}
\usepackage{booktabs} 
\usepackage{tikz}
\usepackage{microtype}
\usepackage{graphicx}
\usepackage{booktabs} 
\usepackage{comment}
\usepackage{nicefrac}
\usepackage{multicol}
\usepackage{multirow}
\usepackage{xspace}
\usepackage[bottom]{footmisc}
\usepackage{bbm}
\usepackage{wrapfig}
\usepackage{threeparttable}
\usepackage{enumitem}
\usepackage{lipsum}
\usepackage[labelfont=bf]{caption}
\usepackage{subcaption}
\usepackage{adjustbox}
\usepackage[group-separator={,}]{siunitx} 

\newif\ifarxiv
\arxivtrue

\usepackage{xcolor}
\definecolor{lightgrey}{rgb}{0.43,0.43,0.43}
\hypersetup{
    colorlinks,
    anchorcolor={blue!50!black},
    linkcolor={blue!50!black},
    citecolor={blue!50!black},
    urlcolor={lightgrey}
}

\usepackage{algorithm}
\usepackage[noend]{algorithmic}

\allowdisplaybreaks 

\newcommand\blfootnote[1]{%
  \begingroup
  \renewcommand\thefootnote{}\footnote{#1}%
  \addtocounter{footnote}{-1}%
  \endgroup
}

\usepackage{Definitions}

\newcommand{\err}[1]{\tiny{$\,$(#1)}}
\newcommand{\highlight}[1]{\textbf{#1}}

\newcommand{\domain}[1]{\textsc{#1}}
\newcommand{\lineardomain}{\domain{Linear}\xspace}
\newcommand{\rffdomain}{\domain{Rff}\xspace}
\newcommand{\grndomain}{\domain{Grn}\xspace}

\newcommand{\obm}{\bm{o}}

\newcommand{\ood}{\text{o.o.d.}\xspace}
\newcommand{\scrna}{{scRNA-seq}\xspace}
\newcommand{\sergio}{{SERGIO}\xspace}
\newcommand{\gnw}{{GeneNetWeaver}\xspace}
\newcommand{\ecoli}{{\em E.\ coli}\xspace}
\newcommand{\yeast}{{\em S.\ cerevisiae}\xspace}
\newcommand\erdosrenyi{\text{Erd{\H{o}}s-R{\'e}nyi}\xspace}

\newcommand{\ges}{\textbf{GES}}
\newcommand{\pc}{\textbf{PC}}
\newcommand{\lingam}{\textbf{LiNGAM}}
\newcommand{\daggnn}{\textbf{DAG-GNN}}
\newcommand{\grandag}{\textbf{GraN-DAG}}
\newcommand{\oursobserv}{\textbf{AVICI} (ours)}

\newcommand{\gies}{\textbf{GIES}}
\newcommand{\igsp}{\textbf{IGSP}}
\newcommand{\dcdi}{\textbf{DCDI}}
\newcommand{\dibs}{\textbf{DiBS}}
\newcommand{\oursinterv}{\textbf{AVICI} (ours)}

\newcommand{\ours}{AVICI\xspace}

\newcommand{\sidratio}{\text{SID}_{\text{ratio}}}

\newcommand{\unif}{\text{Unif}} 
\newcommand{\unifpm}{\text{Unif}_{\pm}}

\usepackage{todonotes}

\newcommand{\vsectiontop}{\vspace{-2pt}}
\newcommand{\vsectionbottom}{\vspace{-2pt}}
\newcommand{\vsubsectiontop}{\vspace{-2pt}}
\newcommand{\vsubsectionbottom}{\vspace{-2pt}}
\newcommand{\vfiguretopofpage}{\vspace{-20pt}}
\newcommand{\vfigurecaptionbelow}{\vspace{-2pt}}
\newcommand{\vspaceabstract}{\vspace{-0pt}}

\title{Amortized Inference for Causal Structure Learning}

\author{
  Lars Lorch\\
  ETH Zurich\\
  Zurich, Switzerland\\
  \href{mailto:llorch@ethz.ch}{\color{black}\texttt{llorch@ethz.ch}}
  \And
  Scott Sussex\\
  ETH Zurich\\
  Zurich, Switzerland\\
  \texttt{ssussex@ethz.ch}\\
  \And
  Jonas Rothfuss\\
  ETH Zurich\\
  Zurich, Switzerland\\
  \texttt{rojonas@ethz.ch}\\
  \AND 
  Andreas Krause${}^*$\\
  ETH Zurich\\
  Zurich, Switzerland\\
  \texttt{krausea@ethz.ch}
  \And
  Bernhard Sch{\"o}lkopf${}^*$\\
  MPI for Intelligent Systems\\
  T{\"u}bingen, Germany\\
  \texttt{bs@tuebingen.mpg.de}
}

\begin{document}

\maketitle

\vspaceabstract
\begin{abstract}
Inferring causal structure poses a combinatorial search problem that typically involves evaluating structures with a score or independence test. The resulting search is costly, and designing suitable scores or tests that capture prior knowledge is difficult. In this work, we propose to {\em amortize causal structure learning}. Rather than searching over structures, we train a variational inference model to directly predict the causal structure from observational or interventional data. This allows our inference model to acquire domain-specific inductive biases for causal discovery solely from data generated by a simulator, bypassing both the hand-engineering of suitable score functions and the search over graphs. The architecture of our inference model emulates permutation invariances that are crucial for statistical efficiency in structure learning, which facilitates generalization to significantly larger problem instances than seen during training. On synthetic data and semisynthetic gene expression data, our models exhibit robust generalization capabilities when subject to substantial distribution shifts and significantly outperform existing algorithms, especially in the challenging genomics domain.
Our code and models are publicly available at: \href{https://github.com/larslorch/avici}{\texttt{https://github.com/larslorch/avici}}.
\end{abstract}

\vsectiontop
\section{Introduction}\label{sec:intro}
\vsectionbottom
\blfootnote{${}^*$Equal supervision.}
\looseness - 1 
Learning the causal structure among a set of variables is a fundamental task in various scientific disciplines  \citep{spirtes2000causation, pearl2009causality}.
However, inferring this causal structure from observations of the variables is a difficult inverse problem.
The solution space of potential causal structures, usually modeled as directed graphs, grows superexponentially with the number of variables. 
To infer a causal structure, standard methods have to search over potential graphs, usually maximizing either a graph scoring function or testing for conditional independences  \citep{heinze2018causal}.

Specifying realistic inductive biases is universally difficult for existing approaches to causal discovery.
Score-based methods use strong assumptions about the data-generating process, such as linearity \citep{shimizu2006linear}, specific noise models \citep{hoyer2008nonlinear,peters2014identifiability}, and the absence of measurement error (cf.~\citealt{scheines2016measurement,zhang2017causal}), which are difficult to verify \citep{dawid2010beware,reisach2021beware}. 
Conversely, constraint-based methods do not have enough domain-specific inductive bias.
Even with an arbitrarily large dataset,
they are limited to identifying equivalence classes that may be exponentially large \citep{he2015counting}.
Moreover, the search over directed graphs itself may introduce unwanted bias and artifacts (cf.~\citealt{colombo2014order}).
The intractable search space ultimately imposes hard constraints on the causal structure, e.g., the node degree \citep{spirtes2000causation}, which limits the suitability of search in real-world domains.

In the present work, we propose to {\em amortize} causal structure learning. 
In other words, our goal is to optimize an inference model to directly predict a causal structure from a provided dataset. 
We show that this approach allows inferring causal structure solely based on synthetic data generated by a {\em simulator} of the data-generating process we are interested in.
Much effort in the sciences, for example, goes into the development of realistic simulators for high-impact and yet challenging causal discovery domains, like
gene regulatory networks \citep{schaffter2011genenetweaver,dibaeinia2020sergio}, 
fMRI brain responses \citep{buxton2009introduction,bassett2017network}, and chemical kinetics \citep{anderson2011continuous,wilkinson2018stochastic}.
Our approach based on amortized variational inference (\ours) ultimately allows us to both specify domain-specific inductive biases not easily represented by graph scoring functions and bypass the problems of structure search. 
Our model architecture is permutation in- and equivariant with respect to the observation and variable dimensions of the provided dataset, respectively, and generalizes to significantly larger problem instances than seen during training.

\looseness-1 

\looseness -1 
On synthetic data and semisynthetic gene expression data, our approach significantly outperforms existing algorithms for causal discovery, often by a large margin. 
Moreover, we demonstrate that our inference models induce calibrated uncertainties and robust behavior when subject to substantial distribution shifts of graphs, mechanisms, noise, and problem sizes.
This suggests that our pretrained models are not only fast but also both reliable and versatile for future downstream use.
In particular, \ours was the only method to infer plausible causal structures from noisy gene expression data, advancing the frontiers of structure discovery in fields such as biology.

\vsectiontop
\section{Background and Related Work}\label{sec:background}
\vsectionbottom

\vsubsectiontop
\subsection{Causal Structure}\label{ssec:background-causal-structure}
\vsubsectionbottom
\looseness-1
In this work, we follow \citet{mooij2016distinguishing} and define the causal structure $G$ of a set of $d$ variables $\xb$~$=$~$(x_1, \dots, x_d)$ as the directed graph over $\xb$ whose edges represent all {\em direct causal} effects among the variables.
A variable $x_i$ has a direct causal effect on $x_j$ if intervening on $x_i$ affects the outcome of $x_j$ independent of the other variables 
$\smash{\xb_{\backslash ij} := \xb\,\backslash\{x_i, x_j\}}$, \ie, there exists $a \neq a'$ such that
\begin{align}\label{eq:causal-effect}
    p(x_j \given \text{do}(x_i = a, \xb_{\backslash ij} = \cb)) \neq p(x_j \given \text{do}(x_i = a', \xb_{\backslash ij} = \cb))
\end{align}
\looseness -1
for some $\cb$.
An \textit{intervention} $\smash{\text{do}(\cdot)}$ denotes any active manipulation of the generative process of $\xb$, like gene knockouts, in which the transcription rates of genes are externally set to zero.
Other models such as causal Bayesian networks and structural causal models \citep{peters2017elements} are less well-suited for describing systems with feedback loops, which we consider practically relevant.
However, we note that our approach does not require any particular formalization of causal structure. In particular, we later show how to apply our approach when $G$ is constrained to be acyclic. 
We assume causal sufficiency, \ie, that $\xb$ contains all common causal parents of the variables $x_i$ \citep{peters2017elements}.

\vsubsectiontop
\subsection{Related Work} \label{ssec:related-work}
\vsubsectionbottom
\looseness-1
Classical methods for causal structure learning search over causal graphs and evaluate them using a likelihood or conditional independence test \citep{chickering2003optimal,kalisch2007estimating,hauser2012characterization,zheng2018dags,heinze2018causal}.
Other methods combine constraint- and score-based ideas \citep{tsamardinos2006max}
or use the noise properties of an SCM that is postulated to underlie the data-generating process \citep{shimizu2006linear,hoyer2008nonlinear}.

\looseness-1
Deep learning has been used for causal inference, \eg, 
for estimating treatment effects \citep{shalit2017estimating,louizos2017causal,yoon2018ganite}
and in instrumental variable analysis \citep{hartford2017deep,bennett2019deep}.
In structure learning, neural networks have primarily been used to model nonlinear causal mechanisms \citep{goudet2018causal,yu2019daggnn,lachapelle2019gradient,brouillard2020differentiable,lorch2021dibs} or to infer the structure of a single dataset \citep{zhu2020causal}.
Prior work applying amortized inference to causal discovery only studied narrowly defined subproblems such as the bivariate case \citep{lopez2015towards} and fixed causal mechanisms \citep{lowe2020amortized} or used correlation coefficients for prediction \citep{li2020supervised}.
In concurrent work, \citet{ke2022learning} also frame causal discovery as supervised learning, but with significant differences.
Most importantly, we optimize a variational objective under a model class that captures the symmetries of structure learning.
Empirically, our models generalize to much larger problem sizes, even on realistic genomics data.

\vsectiontop
\section{AVICI: Amortized Variational Inference for Causal Discovery}
\label{sec:method}
\vsectionbottom

\subsection{Variational Objective}\label{ssec:variational-objective}
\vsubsectionbottom
\looseness-1
To amortize causal structure learning, we define a data-generating distribution $p(D)$ that models the domain in which we infer causal structures.
The observations $D$~$=$~$\{\xb^1, \dots, \xb^n \}$~$\sim$~$p(D)$ are generated by sampling from a distribution over causal structures $p(G)$ and then obtaining realizations from a data-generating mechanism $p(D \given G)$.
The data-generating process $p(D \given G)$ characterizes all direct causal effects (\ref{eq:causal-effect}) in the system, but it is not necessarily induced by ancestral sampling over a directed acyclic graph.
Real-world systems are often more naturally modeled at different granularities or as dynamical systems \citep{mooij2013from,hoel2013quantifying,rubenstein2017causal,scholkopf2019causality}.

Given a set of observations $D$,
our goal is to approximate the posterior over causal structures $p(G \given D)$ with a variational distribution $q(G; \theta)$.
To amortize this inference task for the domain distribution $p(D)$,  we optimize an inference model $f_\phi$ to predict the variational parameters $\theta$ by minimizing the expected {\em forward KL divergence} from the intractable posterior $p(G \given D)$ to $q(G; \theta)$ for $D$~$\sim$~$p(D)$:
\begin{align}\label{eq:forward-kl}
	\min_\phi ~\EE_{p(D)}~  D_{KL} \big( p(G \given D) \big\Vert q\big(G ; f_\phi(D)\big )\big)
\end{align}
Since it is not tractable to compute the true posterior in (\ref{eq:forward-kl}), we make use of ideas by \citet{agakov2004algorithm} and rewrite the expected forward KL to obtain an equivalent, tractable objective:
\begin{align}
\begin{split}\label{eq:tractable-forward-kl}
	\EE_{p(D)}~  D_{KL} \big( p(G \given D) \big\Vert q(G ; f_\phi(D))\big) 
	&=\EE_{p(D)}~\EE_{p(G \given D)} [ \log p(G \given D) - \log q(G ; f_\phi(D)) ] \\
	&= - \EE_{p(G)}~\EE_{p(D \given G)} [ \log q(G ; f_\phi(D)) ] + \text{const.}
\end{split}
\end{align}
The constant does not depend on $\phi$, so we can maximize $\Lcal(\phi) := \EE_{p(G)}~\EE_{p(D \given G)} [ \log q(G; f_\phi(D)) ] $,
which allows us to perform amortized variational inference for causal discovery (\ours).
While the domain distribution $\smash{p(D) = \EE_{p(G)} [ p(D \given G)]}$ can be arbitrarily complex,
$\Lcal$ is tractable whenever we have access to the causal graph $G$ underlying the generative process of $D$, \ie, to samples from the joint distribution $p(G, D)$.
In practice, $p(G)$ and $p(D \given G)$ can thus be specified by a simulator.

From an information-theoretic viewpoint, the objective (\ref{eq:forward-kl}) maximizes a variational lower bound on the mutual information $I[G; D]$ between the causal structure $G$ and the observations $D$ \citep{agakov2004algorithm}.
Starting from the definition of mutual information, we obtain
\begin{align}
\begin{split}
    I[G; D] = H[G] - H[G \given D] 
    &= H[G] + \EE_{p(G, D)} [ \log p(G \given D) ] \\
    &\geq  H[G] + \EE_{p(G, D)} [ \log q(G ; f_\phi(D)) ] 
    = H[G] + \Lcal(\phi)
\end{split}
\end{align}
where the entropy $H[G]$ is constant.
The bound is tight if 
$\EE_{p(D)} D_{KL} ( p(G \given D) \Vert q(G ; f_\phi(D))  ) = 0$.

\vsubsectiontop
\subsection{Likelihood-Free Inference using the Forward KL}\label{ssec:forward-kl}
\vsubsectionbottom
The \ours objective in \eqref{eq:tractable-forward-kl} intentionally targets the forward KL $D_{KL}(p \,\Vert\, q(\,\cdot\,; \theta))$, which requires optimizing
$\smash{\EE_{p(G, D)} [ \log q(G; \theta) ]} $.
This choice implies that we both model the density $q(G; \theta)$ explicitly and assume access to {\em samples}  from the true data-generating distribution $p(G, D)$. Minimizing the forward KL enables us to infer causal structures in arbitrarily complex domains---that is, even domains where it is difficult to specify an explicit likelihood $p(D \given G)$. 
Moreover, the forward KL typically yields more reliable uncertainty estimates since it does not suffer from the variance underestimation problems common to the reverse KL \citep{bishop2006pattern}.

\looseness-1
In contrast, variational inference usually optimizes the reverse KL $D_{KL}(q( \,\Vert\, p)$, which involves the reconstruction term $\smash{\EE_{q(G ; \theta)}[ \log p(D \given G )]}$ \citep{blei2017variational}.
This objective requires a tractable marginal likelihood $p(D \given G)$.
Unless inferring the mechanism parameters jointly
(\eg~\citealt{brouillard2020differentiable,lorch2021dibs}), this requirement limits inference to conjugate models with linear Gaussian or categorical mechanisms that assume zero measurement error
\citep{geiger1994learning,heckerman1995learning}, which are not justified in practice
\citep{friston2000nonlinear,schaffter2011genenetweaver,runge2019inferring,dibaeinia2020sergio}.
Furthermore, unless the noise scale is learned jointly, likelihoods can be sensitive to the measurement scale of $\xb$ \citep{reisach2021beware}.

\vsectiontop
\section{Inference Model}\label{sec:inference-model}
\vsectionbottom
In the following section, we describe a choice for the variational distribution $q(G ; \theta)$ and the inference model $f_\phi$ that predicts $\theta$ given $D$.
After that, we detail our training procedure for optimizing the model parameters $\phi$ and for learning causal graphs with acyclicity constraints.

\subsection{Variational Family}
\vsubsectionbottom
While any inference model that defines a density is feasible for maximizing the objective in (\ref{eq:tractable-forward-kl}), we opt to use a factorized variational family in this work.
\begin{align}
\begin{split} \label{eq:q-model-factorized}
    q(G ; \theta) &= \prod_{i, j} q(g_{i,j} ; \theta_{i,j}) ~~
    \text{with} ~~ g_{i,j} \sim \text{Bern}(\theta_{i,j})
\end{split}
\end{align}
\looseness=-1%
The inference model $f_\phi$ maps a dataset $D$
corresponding to $n$ samples $\smash{\{\obm^{1}, \dots, \obm^{n} \}}$ to a $d$-by-$d$ matrix $\theta$ parameterizing the variational approximation of the causal graph posterior.
In addition to the joint observation $\xb^{i} = (x^i_1, \dots, x^i_d)$,
each sample $\smash{\obm^{i} = (o^{i}_{1}, \dots, o^{i}_{d})}$ may contain interventional information for each variable. 
When interventions or gene knockouts are performed, we set 
$\smash{o^{i}_{j} = (x^{i}_{j}, u^{i}_{j})}$ and $\smash{u^{i}_{j} \in \{0,1\}}$ indicating whether variable $j$ was intervened upon in sample~$i$.
Other settings could be encoded analogously, \eg, when the intervention targets are unknown or measurements incomplete.

\vsubsectiontop
\subsection{Model Architecture} \label{ssec:architecture}
\vsubsectionbottom
\looseness-1
To maximize statistical efficiency, $f_\phi$ should satisfy the symmetries inherent to the task of causal structure learning.
Firstly, $f_\phi$ should be {\em permutation  invariant} across the sample dimension (axis~$n$).
Shuffling the samples should not influence the prediction, \ie,
for any permutation~$\pi$, we have
$f_\phi(\pi(\{\obm\})) = f_\phi(\{\obm\})$.
Moreover,  $f_\phi$ should be {\em permutation equivariant} across the variable dimension (axis~$d$).
Reordering the variables should permute the predicted causal edge probabilities, \ie, $f_\phi(\{\obm_{\pi(1:d)}\})_{i,j} = f_\phi(\{\obm_{1:d} \})_{\pi(i), \pi(j)}$.
Lastly, $f_\phi$ should apply to any $d, n \geq 1$.

\input{fig_architecture}
\looseness-1
In the following, we show how to parameterize $f_\phi$ as a neural network that encodes these properties.
After first mapping each $\smash{o^i_j}$ to a real-valued vector using a position-wise linear layer, $f_\phi$ operates over a continuous, three-dimensional tensor of $n$ rows for the observations, $d$ columns for the variables, and feature size $k$.
Figure \ref{fig:architecture} illustrates the key components of the architecture. 

\looseness-1
\textbf{Attending over axes $d$ and $n$}\quad 
The core of $f_\phi$ is composed of $L = 8$ identical layers.
Each layer consists of four residual sublayers, where the first and third apply multi-head self-attention and the second and fourth position-wise feed-forward networks, similar to the Transformer encoder \citep{vaswani2017attention}.
To enable information flow across all $n \times d$ tokens of the representation, the model alternates in attending over the observation and the variable dimension \citep{kossen2021self}.
Specifically, the first self-attention sublayer attends over axis $d$, treating axis $n$ as a batch dimension; the second attends over axis $n$, treating axis $d$ as a batch dimension.
Since 
modules are shared across non-attended axes, the representation is permutation equivariant over axes $n$ and $d$ at all times \citep{lee2019set}.

\textbf{Variational parameters}\quad 
After building up a representation tensor from the input using the attention layers, we max-pool over the observation axis $n$ to obtain a representation $(\zb^1, \dots, \zb^d)$ consisting of one vector $\zb^i \in \RR^k $ for each causal variable.
Following \citet{lorch2021dibs}, we use two position-wise linear layers to map each $\zb^i$ to two embeddings $\ub^i, \vb^i \in \RR^{k}$, which are $\ell_2$ normalized.
We then model the probability of each edge in the causal graph with an inner product:
\begin{align}\label{eq:dibs}
    \theta_{i,j} = \sigma
    \big(\tau \, \ub^i \cdot \vb^j + b \big)
\end{align}
where $\sigma$ is the logistic function, $b$ a learned bias, and $\tau$ a positive scale that is learned in $\log$ space.
Since max-pooling is invariant to permutations and since (\ref{eq:dibs}) permutes with respect to axis $d$, $f_\phi$ satisfies the required permutation invariance over axis $n$ and permutation equivariance over axis $d$.

\vsubsectiontop
\subsection{Acyclicity}\label{ssec:acyclicity}
\vsubsectionbottom
\looseness-1
Cyclic causal effects often occur, \eg, when modeling stationary distributions of dynamical systems, and thus loops in a causal structure are possible. 
However, certain domains may be more accurately modeled by acyclic structures \citep{rubenstein2017causal}.
While the variational family in (\ref{eq:q-model-factorized}) cannot enforce it, we can optimize for acyclicity through $\phi$.
Whenever the acyclicity prior is justified, we amend the optimization problem in (\ref{eq:forward-kl}) with the constraint that $q$ only models acyclic graphs in expectation:
\begin{align}\label{eq:acyclicity-constraint}
    \Fcal(\phi) := \EE_{p(D)} \left[ h(f_\phi(D)) \right] = 0
\end{align}
The function $h$ is zero if and only if the predicted edge probabilities induce an acyclic graph. 
We use the insight by \citet{lee2019scaling}, who show that acyclicity is equivalent to the spectral radius $\rho$, \ie, the largest absolute eigenvalue, of the predicted matrix being zero.
We use power iteration to approximate and differentiate through the largest eigenvalue of $f_\phi(D)$ \citep{golub2000eigenvalue,lee2019scaling}:
\begin{align}\label{eq:acyclicity-spectral}
    h(W) := \rho(W) \approx \frac{\ab^\top W \bb}{\ab^\top \bb} ~~~\text{where for $t$ steps: }  ~~~ 
    \begin{aligned}
        \ab &\gets \ab^\top W ~/~ \lVert \ab^\top W  \rVert_2 \\
        \bb &\gets W\bb ~/~ \lVert W\bb \rVert_2 
    \end{aligned}
\end{align}
\looseness -1 and $\ab, \bb \in \RR^d$ are initialized randomly. 
Since a few steps $t$ are sufficient in practice, (\ref{eq:acyclicity-spectral}) scales with $O(d^2)$ and is significantly more efficient than  $O(d^3)$ constraints based on matrix powers \citep{zheng2018dags,yu2019daggnn}. 
We do not backpropagate gradients with respect to $\phi$ through $\ab, \bb$.

\vsubsectiontop
\subsection{Optimization}\label{ssec:optimization}
\vsubsectionbottom
\newcommand{\algowidth}{0.47\textwidth} 

\input{fig_algo}

\looseness-1
Combining the objective in (\ref{eq:tractable-forward-kl}) with our inference model (\ref{eq:q-model-factorized}), we can directly use stochastic optimization to train the parameters $\phi$ of the inference model.
The expectations over $p(G, D)$ inside $\Lcal$ and $\Fcal$ are approximated using samples from the data-generating process of the domain. 
When enforcing acyclicity, causal discovery algorithms often use the augmented Lagrangian method for constrained optimization (\eg,~\citealt{zheng2018dags,brouillard2020differentiable}).
In this work, we optimize the parameters $\phi$ of a neural network, so we rely on methods specifically tailored for deep learning and solve the constrained program
$\max_\phi \Lcal(\phi)$
s.t.\ $\Fcal(\phi) = 0$
through its dual formulation \citep{nandwani2019primal}:
\begin{align}
\begin{split}\label{eq:dual-program}
	\min_\lambda \max_\phi~ \Lcal(\phi) - \lambda \Fcal(\phi)
\end{split}
\end{align}
Algorithm \ref{alg:training} summarizes the general optimization procedure for $q_\phi$, which converges to a local optimum under regularity conditions on the learning rates \citep{jin2020what}.
Without an acyclicity constraint, training reduces to the primal updates of $\phi$ with $\lambda = 0$.

\vsectiontop
\section{Experimental Setup}\label{sec:experimental-setup}
\vsectionbottom

\looseness-1
Evaluating causal discovery algorithms is difficult since there are few interesting real-world datasets that come with ground-truth causal structure. 
Often, the believed ground truths may be incomplete or change as expert knowledge improves \citep{schaffter2011genenetweaver,mooij2020joint}.
Following prior work, we deal with this difficulty by evaluating our approach using simulated data with known causal structure and by controlling for various aspects of the task.
In Appendix \ref{app:extended-results}, we additionally report results on a real-world proteomics dataset \citep{sachs2005causal}.

\vsubsectiontop
\subsection{Domains and Simulated Components}\label{ssec:domains}
\vsubsectionbottom
\looseness-1
We study three domains: two classes of structural causal models (SCMs) as well as semisynthetic single-cell expression data of gene regulatory networks (GRNs).
To study the generalization of \ours beyond the training distribution $p(D)$, we carefully construct a spectrum of test distributions $\tilde{p}(D)$ that incur substantial shift from $p(D)$ in terms of the causal structures, mechanisms, and noise, which we study in various combinations.
Whenever we consider interventional data in our experiments, half of the dataset consists of observational data and half of single-variable interventions.

\input{fig_illustration}

\looseness-1
\textbf{Data-generating processes $p(D \given G)$ }\quad
We consider SCMs with linear functions (\lineardomain) and with nonlinear functions of random Fourier features (\rffdomain) that correspond to functions drawn from a Gaussian process with squared exponential kernel \citep{rahimi2007random}. 
In the out-of-distribution (\ood) setting $\tilde{p}(D)$, we sample the linear function and kernel parameters from the tails of $p(D)$ and unseen value ranges.
Moreover, we simulate homoscedastic Gaussian noise in the training distribution $p(D)$ but test on heteroscedastic Cauchy and Laplacian noise \ood that is induced by randomly drawn, nonlinear functions $h_j$. 
In \lineardomain and \rffdomain,
interventions set variables to random values and are performed on a subset of target variables containing half of the nodes.

\looseness-1
In addition to SCMs, we consider the challenging domain of GRNs (\grndomain) using the simulator of \citet{dibaeinia2020sergio}.
Contrary to SCMs, gene expression samples correspond to draws from the steady state of a stochastic dynamical system that varies between cell types \citep{huynh2019gene}.
In the \ood setting, the parameters sampled for the \grndomain simulator are drawn from significantly wider ranges.
In addition, we use the noise levels of different single-cell RNA sequencing technologies, which were calibrated on real datasets. 
In \grndomain, interventions are performed on all nodes and correspond to gene knockouts, forcing the transcription rate of a variable to zero.

\textbf{Causal structures $p(G)$}\quad
\looseness -1 Following prior work, we use random graph models and known biological networks to sample ground-truth causal structures.
In all three domains, the training data distribution $p(D)$ is induced by simple \erdosrenyi and scale-free graphs \citep{erdos1959random,barabasi1999emergence}.
In the \ood setting, $\tilde{p}(D)$ of the \lineardomain and \domain {Rff} domains are simulated using causal structures from
the Watts-Strogatz model, capturing small-world phenomena \citep{watts1998collective};
the stochastic block model, generalizing \erdosrenyi to community structures \citep{holland1983stochastic};
and geometric random graphs, modeling connectivity based on spatial distance \citep{gilbert1961random}.
In the \grndomain domain, we use subgraphs of the known \yeast and \ecoli GRNs and their effect signs whenever known.
To extract these subgraphs, we use the procedure by \citet{marbach2009generating} to maintain structural patterns like motifs and modularity 
\citep{ravasz2002hierarchical,shen2002network}.

\looseness-1
To illustrate the distribution shift from $p(D)$ to $\tilde{p}(D)$, Figure \ref{fig:domain-illustration} shows a set of graph, mechanism, and noise distribution samples in the \rffdomain domain.
In Appendix~\ref{app:domain-spec}, we give the detailed parameter configurations and functions defining $p(D)$ and $\tilde{p}(D)$ in the three domains. We also provide details on the simulator by \citet{dibaeinia2020sergio} and subgraph extraction \citep{marbach2009generating}.

\vsubsectiontop
\subsection{Evaluation Metrics}\label{sec:eval_metrics}
\vsubsectionbottom

\looseness-1
All experiments throughout this paper are conducted on datasets that \ours has never seen during training, regardless of whether we evaluate the predictive performance in-distribution or \ood
To assess how well a predicted structure reflects the ground truth, we report the structural Hamming distance (SHD) and the structural intervention distance (SID) \citep{peters2015structural}. While the SHD simply reflects the graph edit distance, the SID quantifies the closeness of two graphs in terms of their interventional adjustment sets.
For these metrics and for single-edge precision, recall, and F1 score, we convert the posterior probabilities predicted by \ours to hard predictions using a threshold of \num{0.5}. 
We evaluate the uncertainty estimates by computing the areas under the precision-recall curve (AUPRC) and the receiver operating characteristic (AUROC) \citep{friedman2003being}.
How well these uncertainty estimates are calibrated is quantified with the expected calibration error (ECE) \citep{degroot1983comparison}.
More details on the metrics are given in Appendix~\ref{app:metrics}.

\vsubsectiontop
\subsection{Inference Model Configuration}
\vsubsectionbottom

\looseness - 1
We train three inference models overall, one for each domain, and perform all experiments on these three trained models,
both when predicting from only observational and from interventional data.
During training, the datasets sampled from $p(D)$ have $d=2$ to $50$ variables and $n=200$ samples. With probability 0.5, these training datasets contain \num{50} interventional samples.
The inference models in the three domains share identical hyperparameters for the architecture and optimization, except for the dropout rate.
We add the acyclicity constraint for the SCM domains \lineardomain and \rffdomain.
Details on the optimization and architecture are given in Appendix \ref{app:inference-model-config}.

\vsectiontop
\section{Experimental Results}\label{sec:results}
\vsectionbottom

\input{fig_results_generalization}

\subsection{Out-Of-Distribution Generalization}\label{ssec:results-ood}
\vsubsectionbottom

\looseness-1
\textbf{Sensitivity to distribution shift}\quad 
In our first set of experiments, we study the generalization capabilities of our inference models across the spectrum of test distributions described in Section~\ref{ssec:domains}.
We perform causal discovery from $n=1000$ observations in systems of $d=30$ variables.
Starting from the training distribution $p(D)$, we incrementally introduce the described distribution shifts in the causal structures, causal mechanisms, and finally noise, where fully \ood corresponds to $\tilde{p}(D)$.
The top row of Figure \ref{fig:generalization} visualizes the results of an empirical sensitivity analysis. 
The radar plots disentangle how combinations of the three \ood aspects, i.e., graphs, mechanisms, and noise, affect the empirical performance in the three domains \lineardomain, \rffdomain, and \grndomain.
In addition to the metrics in Section \ref{sec:eval_metrics}, we also report the percentage of predicted graphs that are acyclic.

In the \lineardomain domain, \ours performs very well in all metrics and hardly suffers under distribution shift. 
In contrast, \grndomain is the most challenging problem domain and the performance degrades more significantly for the \ood scenarios. 
We observe that \ours can perform better under certain distribution shifts than in-distribution, \eg, in \grndomain. 
This is because \ours empirically performs better at predicting edges adjacent to large-degree nodes, a common feature of the \ecoli and \yeast graphs not present in the \erdosrenyi training structures. 
We also find that acyclicity is perfectly satisfied for \lineardomain and \rffdomain and that AUPRC and AUROC do not suffer as much from distributional shift as the metrics based on thresholded point estimates.

\looseness-1
In Appendix \ref{app:extended-results-generalization}, we additionally report results for generalization from  \lineardomain to \rffdomain and vice versa, \ie, to entirely unseen function classes of causal mechanisms in addition to the previous \ood shifts.

\textbf{Generalization to unseen problem sizes}\quad
\looseness -1 
In addition to the sensitivity to distribution shift, we study the ability to generalize to unseen problem sizes.
The bottom row of Figure \ref{fig:generalization} illustrates the AUPRC for the edge predictions of \ours when varying $d$ and $n$ on unseen in-distribution data. 
The predictions improve with the number of data points $n$ while exhibiting diminishing marginal improvement when seeing additional data. 
Moreover, the performance decreases smoothly as the number of variables $d$ increases and the task becomes harder.
Most importantly, this robust behavior can be observed well beyond the settings used during training ($n=200$ and $d \leq 50$).

\vsubsectiontop
\subsection{Benchmarking}\label{ssec:benchmarking}
\vsubsectionbottom

\input{tab_benchmark_30}

\looseness-1
Next, we benchmark \ours against existing algorithms. 
Using only observational data, we compare with 
the PC algorithm \citep{spirtes2000causation},
GES \citep{chickering2003optimal},
LiNGAM \citep{shimizu2006linear},
DAG-GNN \citep{yu2019daggnn},
and GraN-DAG \citep{lachapelle2019gradient}.
Mixed with interventional data, we compare with 
GIES \citep{hauser2012characterization},
IGSP \citep{wang2017permutation},
and DCDI \citep{brouillard2020differentiable}.
We tune the important hyperparameters of each baseline on held-out task instances of each domain. 
When computing the evaluation metrics, we favor methods that only predict (interventional) Markov equivalence classes by orienting undirected edges correctly when present in the ground truth.
Details on the baselines are given in Appendix \ref{app:baselines}.

\looseness-1
The benchmarking is performed on the fully \ood domain distributions $\tilde{p}(D)$, \ie, under distribution shifts on causal graphs, mechanisms, and noise distributions w.r.t.\ the training distribution of \ours.
Table \ref{tab:benchmark-30} shows the SID and F1 scores of all methods given $n$~$=$~$1000$ observations for $d$~$=$~$30$ variables.
We find that the \ours model trained on \lineardomain outperforms all baselines, both given observational or interventional data, despite operating under significant distribution shift. Only GIES achieves comparable accuracy.
The same holds for \rffdomain, where GraN-DAG and DCDI perform well but ultimately do not reach the accuracy of \ours.

\looseness-1
In the \grndomain domain, where inductive biases are most difficult to specify, classical methods fail to infer plausible graphs.
However, provided interventional data,
\ours can use its learned inductive bias to infer plausible causal structures from the noisy gene expressions, even under distribution shift. This is a promising step towards reliable structure discovery in fields like molecular biology.
Even without gene knockout data, \ours achieves nontrival AUROC and AUPRC while classical methods predict close to randomly (Table \ref{tab:benchmark-30-prob} in Appendix \ref{app:extended-results}; see also \citealt{dibaeinia2020sergio,chen2018evaluating}).

Results for in-distribution data and for larger graphs of $d$~$=$~$100$ variables are given in Appendices \ref{app:extended-results-30-in-distribution} and \ref{app:extended-results-100}. 
In Appendix \ref{app:extended-results-sachs}, we also report results for a real proteomics dataset \citep{sachs2005causal}.

\input{fig_calibration_combined}

\textbf{Uncertainty quantification}\quad
\looseness-1
Using metrics of calibration, we can evaluate the degree to which predicted edge probabilities are consistent with empirical edge frequencies \citep{degroot1983comparison,guo2017calibration}.
We say that a predicted probability $\smash{p}$ is calibrated if we empirically observe an event in $\smash{(p\cdot100)\%}$ of the cases.
When plotting the observed edge frequencies against their predicted probabilities, a calibrated algorithm induces a diagonal line.
The expected calibration error (ECE) represents the weighted average deviation from this diagonal.
For further details, see Appendix \ref{app:metrics}.

Since the baseline algorithms only infer point estimates of the causal structure, we use the nonparametric DAG bootstrap to estimate edge probabilities (\citealt{friedman1999data}, Appendix \ref{app:baselines}).
We additionally compare \ours with DiBS, which infers Bayesian posterior edge probabilities like \ours \citep{lorch2021dibs}.
Figure \ref{fig:calibration-avici} gives the calibration plots for \ours and Table \ref{fig:calibration-avici}b the ECE for all methods. 
In each domain, the marginal edge probabilities predicted by \ours are the most calibrated in terms of ECE.
Moreover, Figure \ref{fig:calibration-avici}a shows that \ours closely traces the perfect calibration line, which highlights its accurate uncertainty calibration across the probability spectrum.

In Appendix \ref{app:extended-results-calibration}, we additionally report AUROC and AUPRC metrics for all methods.
We also provide calibration plots analogous to Figure \ref{fig:calibration-avici} for the baselines (Figure \ref{fig:calibration-all}), which often show vastly overconfident predictions where the calibration line is far below the diagonal.

\vsubsectiontop
\subsection{Ablations}\label{ssec:ablations}
\vsubsectionbottom
\looseness-1
Finally, we analyze the importance of key architecture components of the inference network $\smash{f_\phi}$.
Focusing on the \rffdomain domain, we train several additional models and ablate single architecture components. 
We vary the network depth $L$, the axes of attention, the representation of $\theta$, and the number of training steps for $\phi$. All other aspects of the model, training and data simulation remain unchanged.

Table \ref{tab:ablation} summarizes the results.
Most noticeably, we find that the performance drops significantly when
attending only over axis $d$ and aggregating information over axis $n$ only once through pooling after the $2L$ self-attention layers.
Attending only over axis $n$ is not sensible since variable interactions are not processed until the prediction of $\theta$, but we still include the results for completeness.

We also test an alternative variational parameter model given by $\smash{\theta_{i,j} = \phi^\top_\theta \tanh \left (\phi_u^\top \mathbf{u}^i + \phi_v^\top \mathbf{v}^j \right)}$ that uses an additional, learned vector $\smash{\phi_\theta}$ and matrices $\smash{\phi_u, \phi_v}$.
This model has been used in related causal discovery work for searching over high-scoring causal DAGs \citep{zhu2020causal} and is a relational network \citep{santoro2017simple}.
This variant also satisfies permutation equivariance (cf.\ Section \ref{ssec:architecture}) since it applies the same MLP elementwise to each edge pair $[\mathbf{u}^i, \mathbf{v}^j]$.
Ultimately, we find no statistically significant difference in performance to our simpler model in Eq.\ (\ref{eq:dibs}), hence we opt for less parameters and a lower memory requirement.

\looseness-1
Lastly, Table \ref{tab:ablation} shows that the causal discovery performance of \ours scales up monotonically with respect to network depth and training time.
Even substantially smaller models of $L$~$=$~$4$ or shorter training times achieve an accuracy that is on par with most baselines (cf.~Table \ref{tab:benchmark-30}).
Our main models ($\star$) have a moderate size of $4.2$~$\times$~$10^6$ parameters, which amounts to only \num{17.0} MB at f\num{32} precision. 
Performing causal discovery (computing a forward pass) given on a trained model takes only a few seconds on CPU.

\input{tab_ablation}

\vsectiontop
\section{Discussion}
\vsectionbottom
\looseness-1
We proposed \ours, a method for inferring causal structure by performing amortized variational inference over an arbitrary data-generating distribution. 
Our approach leverages the insight that inductive biases crucial for statistical efficiency in structure learning might be more easily encoded in a simulator than in an inference technique. 
This is reflected in our experiments, where \ours solves structure learning problems in complex domains intractable for existing approaches \citep{dibaeinia2020sergio}.
Our method can likely be extended to other typically difficult domains, including settings where we cannot assume causal sufficiency \citep{bhattacharya2021differentiable}.
Our approach will continually benefit from ongoing efforts in developing (conditional) generative models and domain simulators.

\looseness-1
Using \ours still comes with several trade-offs. 
First, while optimizing the dual program empirically induces acyclicity, this constraint is not satisfied with certainty using the variational family considered here. 
Moreover, similar to most amortization techniques \citep{amos2022tutorial}, \ours gives no theoretical guarantees of performance.
Some classical methods can do so in the infinite sample limit given specific assumptions on the data-generating process \citep{peters2017elements}.
However, future work might obtain guarantees for \ours that are similar to learning theory results for the bivariate causal discovery case 
\citep{lopez2015towards}.

Our experiments demonstrate that our inference models are highly robust to distributional shift, suggesting that the trained models could be useful out-of-the-box in causal structure learning tasks outside the domains studied in this paper.
In this context, fine-tuning a pretrained \ours model on labeled real-world datasets is a promising avenue for future work.
To facilitate this, our code and models are publicly available at: \href{https://github.com/larslorch/avici}{\texttt{https://github.com/larslorch/avici}}.

\vsectiontop
\begin{ack}
\vsectionbottom
\looseness-1
We thank Alexander Neitz, Giambattista Parascandolo, and Frederik Tr{\"a}uble for their feedback and the reviewers for their helpful comments.
This research was supported by the European Research Council (ERC) under the European Union's Horizon 2020 research and innovation program grant agreement no.\ 815943 and the Swiss National Science Foundation under NCCR Automation, grant agreement 51NF40 180545.
Jonas Rothfuss was supported by an Apple Scholars in AI/ML fellowship.
\end{ack}

{
\small
\bibliographystyle{apalike}  
\bibliography{ref.bib}
}

\newpage

\newpage
\clearpage

\appendix


\section{Domain Specification and Simulation}\label{app:domain-spec}

In this section, we define the training and test distributions $p(D)$ and $\tilde{p}(D)$ concretely in terms of the parameters and notation introduced in the main text. 
Based on these definitions,
Table \ref{tab:domain_spec_linear_rff} summarizes all parameters of the data-generating processes for \lineardomain and \rffdomain and specifies how they are sampled for a random task instance. 
Table \ref{tab:domain_spec_gene} lists the same specifications for the \grndomain domain.
The notation and parameters are defined in the following subsections.

\input{tab_domain_spec_rff}

\input{tab_domain_spec_gene}

\clearpage
\newpage

\subsection{Causal Structures}

\subsubsection{Random graph models}
In \erdosrenyi graphs, each edge is sampled independently with a fixed probability
\citep{erdos1959random}.
We scale this probability to obtain $O(d)$ edges in expectation.
Scale-free graphs are generated by a sequential preferential attachment process, where in- or outgoing edges of node $i$ to the previous $i-1$ nodes are sampled with probability~$\propto$~$\text{deg}(j)^\alpha$
\citep{barabasi1999emergence}.
Watts-Strogatz graphs are $k$-dimensional lattices, whose edges get rewired globally to random nodes with a specified probability \citep{watts1998collective}.
The stochastic block model generalizes \erdosrenyi to capture community structure. 
Splitting the nodes into a random partition of so-called blocks, the inter-block edge probability is dampened by a multiplying factor compared to the intra-block probability, also tuned to result in $O(d)$ edges in expectation \citep{holland1983stochastic}.
Lastly, geometric random graphs model connectivity based on two-dimensional Euclidian distance within some radius, where nodes are randomly placed inside the unit square \citep{gilbert1961random}.

For undirected random graph models, we orient edges by selecting the upper-triangular half of the adjacency matrix.
The classes of random graph models are sampled in equal proportion when generating a set of evaluation datasets (Tables \ref{tab:domain_spec_linear_rff} and \ref{tab:domain_spec_gene}).

\subsubsection{Subgraph Extraction from Real-World Networks}\label{app:subgraph}
For the evaluation in the \grndomain domain, we sample realistic causal graphs by extracting subgraphs from the known \ecoli and \yeast regulatory networks.
For this, we rely on the procedure by \citet{marbach2009generating}, which is also used by \citet{schaffter2011genenetweaver} and \citet{dibaeinia2020sergio}.
Their graph extraction method is carefully designed to capture the structural properties of biological networks by preserving the functional and structural properties of the source network.

\paragraph{Algorithm}
\looseness -1 The procedure extracts a random subgraph of the source network by selecting a subset of nodes $\Vcal$, and then returning the graph containing all edges from the source network covered by $\Vcal$.
Starting from a random seed node, the algorithm proceeds by iteratively adding new nodes to $\Vcal$.
In each step, this new node is selected from the set of neighbors of the current set $\Vcal$.
The neighbor to be added is selected greedily such that the resulting subgraph has maximum modularity \citep{marbach2009generating}.

To introduce additional randomness, \citet{marbach2009generating} propose to randomly draw the new node from the set of neighbors inducing the top-$p$ percent of the most modular graphs.
In our experiments, we adopt the latter with $p=20$ percent, similar to \citet{schaffter2011genenetweaver}.
The original method of \citet{marbach2009generating} is intended for undirected graphs.
Thus, we use the undirected skeleton of the source network for the required modularity and neighborhood computation.

\paragraph{Real GRNs}
We take the \ecoli and \yeast regulatory networks as provided by the \gnw repository,%
\footnote{\href{https://github.com/tschaffter/genenetweaver}{\texttt{https://github.com/tschaffter/genenetweaver}}} which have \num{1565} and \num{4441} nodes (genes), respectively.
For \ecoli, we also know the true signs of a large proportion of causal effects.
When extracting a random subgraph from \ecoli, we take the true signs of the effects and map them onto the randomly sampled interaction terms $k \in \RR^{d \times d}$ used by \sergio; cf.\ Section \ref{app:sergio}.
When the interaction signs are unknown or uncertain in \ecoli, we impute a random sign in the interaction terms $k$ of \sergio based on the frequency of known positive and negative signs in the \ecoli graph.

Empirically, individual genes in \ecoli tend to predominantly have either up- or down-regulating effects on their causal children. To capture this aspect in \yeast also, we fit the probability of an up-regulating effect caused by a given gene in \ecoli to a Beta distribution.
For each node $j$ in an extracted subgraph of \yeast, we draw a probability $p_j$ from this Beta distribution and then sample the effect signs for the outgoing edges of node $j$ using $p_j$.
As a result, the genes in the subgraphs of \yeast individually also have mostly up- or down-regulating effects. 
Maximum likelihood estimation for this Beta distribution yielded $\alpha = 0.2588$ and $\beta = 0.2499$.

The \ecoli and \yeast graphs and effect signs used in the experiments are taken from the GeneNetWeaver repository \citep{schaffter2011genenetweaver} (MIT License).

\subsection{Data-Generating Processes}

\subsubsection{Structural Causal Models}
In the \lineardomain and \rffdomain domains, the data-generating processes are modeled by structural causal models (SCMs).
In this work, we consider SCMs with causal mechanisms that model each causal variable $x_j$ given its parents $\xb_{\text{pa}(j)}$ as
\begin{align}
    x_j \gets f_j(\xb_{\text{pa}(j)}, \epsilon_j) = f_j(\xb_{\text{pa}(j)}) + h_j(\xb_{\text{pa}(j)})\epsilon_j
\end{align}
where the noise $\epsilon_j$ is additive and may be heteroscedastic through an input-dependent noise scale $\smash{h_j(\xb_{\text{pa}(j)})}$.
Even in the homogeneous noise setting, the scale of each noise distribution $p(\epsilon_j)$ is random and thus different for each variable $x_j$.
We write $\xb_{\text{pa}(j)}$ when indexing $\xb$ at the parents of node $j$.
In the heteroscedastic setting, we parameterize the noise scales as $h_j(\xb) = \log(1+\exp(g_j(\xb))$ for a set of nonlinear functions $g_j$. 

Prior to performing inference with \ours or any baseline, each set of SCM observations $D$ is standardized variable-wise by subtracting its mean and dividing by its standard deviation, so that each $x_j$ has mean \num{0} and variance \num{1}, avoiding potential varsortability bias \citep{reisach2021beware}. 

In the \lineardomain domain, the functions $f_j$ are given by affine transforms
\begin{align}
    f_j(\xb_{\text{pa}(j)}) = \wb_j^\top \xb_{\text{pa}(j)} + b_j
\end{align}
whose weights $\wb_j$ and bias $b_j$ are sampled independently for each $f_j$.
In the \rffdomain domain, the functions $f_j$ modeling each causal variable $x_j$ given its parents $\xb_{\text{pa}(j)}$ are drawn from a Gaussian Process
\begin{align}
    f_j
    \sim  \mathcal{GP}(b_j, k_j)
\end{align}
with bias $b_j$ and squared exponential (SE) kernel $k_j(\xb, \xb') = c_j^2 \exp \left( - \lVert \xb-\xb'\rVert_2^2 / 2\ell_j^2 \right)$ with length scale $\ell_j$ and output scale $c_j$.
The parameters $b_j$, $c_j$, and $\ell_j$ are sampled independently for each variable $j$.
To obtain explicit function draws $f_j$ from the GP, we approximate $f_j$ with random Fourier features  \citep{rahimi2007random}.
Specifically, we can obtain $f_j \sim \mathcal{GP}(b_j, k(\xb, \xb'))$ for a SE kernel $k$ with length scale $\ell_j$ and output scale $c_j$ by sampling
\begin{align}
    f_j(\xb_{\text{pa}(j)}) = b_j + c_j\,{\sqrt{\tfrac{2}{M}}}\, \sum_{m=1}^{M} \alpha^{(m)} \cos \left ( \tfrac{1}{\ell_j}\, \bm{\omega}^{(m)} \cdot \xb_{\text{pa}(j)} + \delta^{(m)} \right)
\end{align}
with 
$\alpha^{(m)} \sim  \Ncal(0, 1)$,
$\bm{\omega}^{(m)} \sim \Ncal(0, \Ib)$,
and
$\delta^{(m)} \sim \unif(0, 2\pi)$.
Throughout this work, we use $M$~$=$~$100$. The function draws become faithful GP samples as $M\rightarrow \infty$ \citep{rahimi2007random}. 
When $x_j$ is a root node and thus has no parents, $f_j$ is a constant.

\subsubsection{Single-Cell Gene Expression Data }\label{app:sergio}
In the \grndomain domain, our goal is to evaluate causal discovery from realistic gene expression data.
There exist several models to simulate the mechanisms, intervention types, and technical measurement noise underlying single-cell expression data of gene regulatory networks
\citep{schaffter2011genenetweaver,huynh2019gene,dibaeinia2020sergio}. 
We use the simulator by \citet{dibaeinia2020sergio} (\sergio) because it resembles the data collected by modern high-throughput single-cell RNA sequencing (\scrna) technologies.
Related genomics simulators, for example, GeneNetWeaver \citep{schaffter2011genenetweaver}, were developed for the simulation of microarray gene expression platforms.
In the following, we give an overview of how to simulate \scrna data with \sergio.
\citet{dibaeinia2020sergio} provide all the details and additional background from the related literature.

\paragraph{Simulation}
Given a causal graph over $d$ genes and a specification of the simulation parameters,
\sergio generates a synthetic scRNA-seq dataset $D$ in two stages.
The $n$ observations in $D$ correspond to $n$ cell samples, that is, the expressions of the $d$ genes recorded in a single cell corresponds to one row in $D$.

In the first stage, \sergio simulates clean gene expressions by sampling randomly-timed snapshots from the steady state of a dynamical system.
In this regulatory process, the genes are expressed at rates influenced by other genes using the chemical Langevin equation, similar to \citet{schaffter2011genenetweaver} and \citep{dibaeinia2020sergio}.
The source nodes in the causal graph $G$ are denoted master regulators (MRs), whose expressions evolve at constant production and decay rates.
The expressions of all downstream genes evolve nonlinearly under production rates caused by the expression of their causal parents in $G$.
Cell {\em types} are defined by specifications of the MR production rates, which significantly influence the evolution of the system.
Thus, the dataset contains variation due to biological system noise within collections of cells of the same type and due to different cell types.
Ultimately, we generate single-cell samples collected from five to ten cell types \citep{dibaeinia2020sergio}.

In the second stage, the clean gene expressions sampled previously are corrupted with technical measurement error that resembles the noise phenomena found in real \scrna data:
\begin{itemize}[leftmargin=20pt]
    \item {\em outlier genes}: a small set of genes have unusually high expression across measurements
    \item {\em library size}: different cells have different total UMI counts, following a log-normal distribution
    \item {\em dropouts}: a high percentage of genes are recorded with zero expression in a given measurement
    \item {\em unique molecule identifier (UMI) counts}: we observe Poisson-distributed count data rather than the clean expression values
\end{itemize}
To configure these noise modules, we use the parameters calibrated by \citet{dibaeinia2020sergio} for datasets from different \scrna technologies.
We extend \sergio to allow for the generation of knockout intervention experiments.
For this, we force the production rate of knocked-out genes to zero during simulation.
Our implementation uses the public source code by \citep{dibaeinia2020sergio}, which is available under a GNU General Public License v3.0.\footnote{\href{https://github.com/PayamDiba/SERGIO}{\texttt{https://github.com/PayamDiba/SERGIO}}}

\paragraph{Parameters}
Given a causal graph $G$, the parameters \sergio requires to simulate $c$ cell types of $d$ genes are:
\begin{itemize}[leftmargin=20pt]
    \item $k \in \RR^{d \times d}$: interaction strengths (only used if edge $i\rightarrow j$ exists in $G$)
    \item $b \in \RR_+^{d \times c}$: MR production rates (only used if gene $j$ is a source node in $G$)
    \item $\gamma \in \RR_+^{d \times d}$: Hill function coefficients controlling nonlinearity of interactions
    
    \item $\lambda \in \RR^{d}$: decay rates per gene
    
    \item $\zeta \in \RR^{d}_+$: scales of stochastic process noise per gene for chemical Langevin equations
    
\end{itemize}
The technical noise components are configured by:
\begin{itemize}[leftmargin=20pt]
    \item $p_{\text{outlier}} \in [0,1]$: probability that a gene is an outlier gene 
    \item $\mu_{\text{outlier}} \in \RR_+, \sigma_{\text{outlier}} \in \RR_+$: parameters of the log-normal distribution for the outlier multipliers
    
    \item $\mu_{\text{lib}} \in \RR_+, \sigma_{\text{lib}} \in \RR_+$: parameters of the log-normal distribution for the library size multipliers
        
    \item $\delta \in [0,100], \xi \in \RR_+$: dropout percentile and temperature of the logistic function parameterizing the dropout probability of a recorded expression
\end{itemize}
In our experiments, the simulator parameters are selected in the ranges suggested by \citet{dibaeinia2020sergio}.

\paragraph{Standardization}
There are several ways to preprocess and normalize single-cell transcriptomic data for downstream use \citep{robinson2010edger}. 
For simplicity, we employ $\log_2$ counts-per-million (CPM) normalization, which normalizes the total UMI counts per sample and then $\log_2$-transforms the relative count values.
Specifically, the CPM value for gene $j$ in sample $i$ is defined as 
\begin{align}
   x^{\text{cpm}, i}_{j} := \frac{x^{i}_{j} \cdot 10^6}{l_i} \quad \text{with library size } l_i = \sum_{j=1}^d x^{i}_{j} \, \text{.}
\end{align}
For zero expressions $\smash{x_j^i}$, the $\log_2$-CPM values are imputed with zero.
The remaining $\log_2$-CPM values range between \num{10} and \num{19}, so we shift and scale the values before performing causal discovery. 
To replicate the sparsity pattern and the relative ordering of values within samples in the original dataset $D$, we standardize the nonzero $\log_2$-CPM values by subtracting the minimum (instead of the mean) and dividing by the overall standard deviation. 
All methods considered in Section \ref{sec:results}, including \ours, work with \grndomain data in this standardized $\log_2$-CPM format.

\section{Evaluation Metrics}\label{app:metrics}

We report several metrics to assess how well the predicted causal structures reflect the ground-truth graph. 
We measure the overall accuracy of the predictions and how well-calibrated the estimated uncertainties in the edge predictions are, since \ours predicts marginal probabilities $q(g_{i,j}; \theta_{i,j})$ for every edge. 
Unless evaluating these edge probabilities, we use a decision threshold of \num{0.5} to convert the \ours prediction to a single graph $G$.

\paragraph{Structural and edge accuracy}
The structural hamming distance (SHD) \citep{tsamardinos2006max}
reflects the graph edit distance between two graphs, i.e., the edge changes required to transform $G$ into $G'$.
By contrast, the structural intervention distance (SID) \citep{peters2015structural} quantifies the closeness of two DAGs in terms of their valid adjustment sets, which more closely resembles our intentions of using the inferred graph for downstream causal inference tasks.

SHD and SID capture global and structural similarity to the ground truth, but notions like precision and recall at the edge level are not captured well.
SID is zero if and only if the true DAG is a subgraph of the predicted graph, which can reward dense predictions (Prop.\ 8 by \citet{peters2015structural}: SID$(G, G')=0$ when $G$ is empty and $G'$ is fully connected).
Conversely, the trivial prediction of an empty graph achieves highly competitive SHD scores for sparse graphs.

For this reason, we report additional metrics that quantify both the trade-off between precision and recall of edges as well as the calibration of their uncertainty estimates.
Specifically, given the binary predictions for all $d^2$ possible edges in the graph $G$, we compute the edge precision, edge recall, and their harmonic mean (F1-score) for each test case and estimate their means and standard errors across the test cases.
Since the F1-score is high only when precision and recall are high, both empty and dense predictions are penalized and no trivial prediction scores well, making it a reliable metric for structure learning.

\paragraph{Edge confidence}
To evaluate the edge probabilities predicted by \ours and the baselines, we compute the areas under the precision-recall curve (AUPRC) and receiver operating characteristic (AUROC) when converting the probabilities into binary predictions using varying decision thresholds \citep{friedman2003being}.
Both statistics capture different aspects of the confidence estimates.
The AUROC is insensitive to changes in class imbalance (edge vs.\ no-edge) for a given $d$.
However, when the number of variables $d$ in sparse graphs of $O(d)$ edges increases, AUROC increasingly discounts the accuracy on the shrinking proportion of edges present in the ground truth, which makes AUPRC more suitable for comparisons ranging over different $d$.
The AUROC is equivalent to the probability that the method ranks a randomly chosen positive instance (i.e., an edge $i \rightarrow j$ present in the ground truth) higher than a randomly chosen negative instance (i.e., an edge $i \dots j$ absent in the ground truth) \citep{fawcett2004roc}.

\paragraph{Calibration}
To assess the true correctness likelihood implied by the predicted edge probabilities, we use the concept of calibration \citep{degroot1983comparison,guo2017calibration}.
A classifier is said to be calibrated if a predicted edge probability of $\hat{p}_{i,j}$ empirically results in the observation of an edge in $(\hat{p}_{i,j} \times 100)$\% of the cases, \ie,
\begin{align}\label{eq:calibration-def}
    \mathbb{P}\big(g_{i,j} = 1 \given \hat{p}_{i,j} = p\big) = p \, .
\end{align}
Following \citet{guo2017calibration}, we can estimate the degree to which this property is satisfied for the predicted probabilities by defining $M$ intervals $I_m = (\tfrac{m-1}{M}, \tfrac{m}{M})$ and
binning all instances $i,j$ where $\hat{p}_{i,j} \in I_m$ into a set $S_m$.
The empirical confidence and accuracy per bin $S_m$ are then defined as 
\begin{align}
    \text{predicted}~\hat{p}(S_m) = \frac{1}{|S_m|} \sum_{{i,j} \in S_m} \hat{p}_{i,j}~~~~~~~~~~~~
    \text{empirical}~p(S_m) = \frac{1}{|S_m|} \sum_{{i,j} \in S_m} g_{i,j}
\end{align}
where a calibrated classifier has $\text{predicted}~\hat{p}(S_m) = \text{empirical}~p(S_m)$, analogous to (\ref{eq:calibration-def}).
Thus, a calibrated edge classifier induces a diagonal line when plotting the empirical $p(S_m)$ against the predicted $\hat{p}(S_m)$.
The expected calibration error (ECE) is a scalar summary of this calibration plot and amounts to the weighted average of the vertical deviation from the perfect calibration line, \ie,
\begin{align}\label{eq:calibration-ece}
    \text{ECE} = \sum_{m=1}^M \frac{|S_m|}{n} \left |\text{empirical}~p(S_m) - \text{predicted}~\hat{p}(S_m) \right |
\end{align}
where $n$ is the total number of evaluated samples (\ie, edges).
The ECE does not capture accuracy in the sense of being able to predict all classes with high certainty, for which the other metrics are more suitable, but rather whether predicted probabilities are reflective empirical likelihood \citep{guo2017calibration}.
In this work, we use $M=10$ bins to compute the calibration plot lines and the ECE.
The plotted calibration lines compute the calibration statistics in aggregate over all test cases to reduce the variance of the empirical counts within the bins, thus not showing standard errors.

\section{Inference Model Details}\label{app:inference-model-config}

\subsection{Optimization}

\paragraph{Batch sizes}
Each \ours model is trained as described in Algorithm \ref{alg:training}.
The objective $\Lcal(\phi)$ relies on samples from the domain distribution $p(G, D)$ to perform Monte Carlo estimation of the expectations.
During training, the number of variables $d$ in the simulated systems are chosen randomly from
\begin{align}
    d \in \{2, 5, 10, 20, 30, 40, 50 \}
\end{align}
The datasets $D$ in the training distributions always have $n=200$ samples, where with probability \num{0.5}, the observations in a given dataset contain \num{50} interventional samples.
The dimensionality of these training instances $G, D$ varies significantly with the number of variables $d$ 
and, therefore, so do the memory requirements of the forward passes of the  inference model~$f_\phi$.

Given these differences in problem size, we make efficient use of the GPU resources during training by performing individual primal updates in Algorithm \ref{alg:training} using only training instances $(G, D)$ with exactly $d$ variables, where $d$ is randomly sampled in each update step. 
Fixing the number of observations to $n=200$, this allows us to increase the batch size for each considered $d$ to the maximum possible given the available GPU memory 
(in our case ranging from batch sizes of \num{27} for $d=2$ down to \num{6} for $d=50$, per \num{24} GiB GPU device).

During training, we tune the sampling probability of a given $d$ to ensure that $f_\phi$ sees roughly the same number of training data sets for each $d$, \ie, we oversample higher $d$, for which the effective batch size per update step is smaller.
We also scale $\Lcal(\phi)$ by dividing by $d^2$ to ensure an approximately equal loss and hence gradient scale across the different $d$ seen at training time.

The penalty $\Fcal(\phi)$ for the acyclicity constraint is estimated using the same minibatch as for $\Lcal(\phi)$.

\paragraph{Buffer}

Since we have access to the complete data-generating process rather than only a fixed dataset, we approximate $\Lcal(\phi)$ with minibatches that are sampled uniformly randomly from a buffer, which is continually updated with fresh data from $p(G, D)$.
Specifically, we initialize a first-in-first-out buffer that holds \num{200} pairs $(G, D)$ for each unique number of variables $d$ considered during training.
A pool of asynchronous single-CPU workers then constantly generates novel training data and replaces the oldest instances in the buffer using a producer-consumer workflow.
We implement this buffer using an Apache PyArrow Plasma object store (Apache Licence 2.0).
During training, we used \num{128} CPU workers
(Appendix \ref{app:extended-results}).

The workers balance the data generation for different buffers to ensure an equal sample-to-insert ratio across $d$, 
accounting for the oversampling of higher $d$ as well as the longer computation time needed for generating data $D$ of larger $d$, for instance, in the \grndomain domain.
In addition, the dataset $D$ of each element $(G, D)$ in the buffer contains four times more observations than $n=200$ used during training.
These observations are subsampled to obtain $n=200$ each time a given buffer element $(G, D)$ is drawn to introduce additional diversity in the training data in case buffer elements are sampled more than once.

\paragraph{Parameter updates}
The primal updates of the inference model parameters $\phi$ are performed using the LAMB optimizer with a constant base learning rate $3\,{\cdot}\,10^{-5}$ and adaptive square-root scaling by the maximum effective batch size\footnote{With 8 GPU devices, this corresponds to a learning rate of $3\,{\cdot}\,10^{-5}\,{\cdot}\,\sqrt{8\,{\cdot}\,27} \approx 4.4\,{\cdot}\,10^{-4}$ \citep{you2019large}}\citep{you2019large}.
Gradients with respect to $\phi$ are clipped at a global $\ell^2$ norm of one \citep{pascanu2013difficulty}.
In all three domains, we optimize $\phi$ for a total number of \num{300000} primal steps, reducing the learning rate by a factor of ten after \num{200000} steps.

When adding the acyclicity contraint in \lineardomain and \rffdomain, we use a dual learning rate of $\eta = 10^{-4}$ and perform a dual update every \num{500} primal steps.
The dual learning rate $\eta$ is warmed up with a linear schedule from zero over the first \num{50000} primal steps.
To reduce the variance in the dual update, we use an exponential moving average of $\Fcal(\phi)$ with step size $10^{-4}$ maintained during the updates of the primal objective.
To approximate the spectral radius in Eq.~(\ref{eq:acyclicity-spectral}), we perform $t=10$ power iterations initialized at $\ub, \vb \sim \Ncal(0, \Ib_d)$.

\subsection{Architecture}
As described in Section \ref{ssec:architecture}, the core of our model consists of $L=8$ layers, each containing four residual sublayers.
Different from the vanilla Transformer encoder, we employ layer normalization before each multi-head attention and feedforward module and after the last of the $L$ layers \citep{radford2019language}.
The multi-head attention modules have a model size of \num{128}, key size of \num{32}, and \num{8} attention heads.
The feedforward modules have a hidden size of \num{512} and use ReLU activations.
In held-out tasks of \rffdomain and \grndomain, we found that dropout in the Transformer encoder does not hurt performance in-distribution, so we increased the dropout rates from \num{0.0} to \num{0.1} and \num{0.3}, respectively, to help generalization \ood
Dropout, when performed, is done before the residual layers are added, as in the vanilla Transformer \citep{vaswani2017attention}.

The position-wise linear layers that map the two-dimensional representation $(\zb^1, \dots, \zb^d)\in \RR^{d \times k}$ to $\ub^i$ and $\vb^i$, respectively, apply layer normalization prior to their transformations.
We use Kaiming uniform initialization for the weights \citep{he2015delving}.
The bias term inside the logistic function of Eq.~(\ref{eq:dibs}) is initialized at \num{-3} and learned alongside all other parameters $\phi$.
Likewise, the scale parameter $\tau$ is learned but optimized in $\log$ space to ensure positivity, \ie, $\tau = \exp(\tau_{\log})$ where $\tau_{\log}$ is updated as part of $\phi$ and initialized at \num{2}.
When optimizing models under the acyclicity constraint, we ignore the diagonal predictions $\theta_{ii}$ and mask the corresponding loss terms.

\medskip 

We implement \ours with Haiku in JAX \citep{haiku2020github,jax2018github}.
We converged to the above optimization and architecture specifications through experimentation on held-out instances from the training distributions $p(D)$, i.e., in-distribution.

\section{Baselines} \label{app:baselines}

\paragraph{Algorithms and hyperparameter tuning}
We calibrate important hyperparameters for all methods on held-out problem instances from the test data distributions $\tilde{p}(D)$ of \lineardomain, \rffdomain and \grndomain, individually in each domain.
For the following algorithms, we search over the parameters relevant for controlling sparsity and the complexity of variable interactions:
\begin{itemize}[leftmargin=20pt]

    \item DCDI \citep{brouillard2020differentiable}: sparsity regularizer $\lambda \in \{10^{-2}, 10^{-1}, 1 \}$, size of hidden layer in MLPs modeling the conditional distributions $\in \{8, 32\}$  
    
    \item DAG-GNN \citep{yu2019daggnn}: graph thresholding parameter $\in \{0.1, 0.2, 0.3\}$, size of hidden layer in MLP encoder and decoder $\in \{16, 64\}$
    
    \item DiBS \citep{lorch2021dibs}: latent kernel length scale $\ell_z \in \{3, 10, 30\}$, \\
    \dots with BGe marginal likelihood (\lineardomain): effective sample size (sparsity) $\smash{\alpha_\mu^\text{BGe}} \in \{0.1, 1.0\}$\\
    \dots with nonlinear Gaussian likelihood (\rffdomain,~\grndomain): parameter length scale $\ell_\theta \in \{30, 300, 3000\}$

    \item GraN-DAG \citep{lachapelle2019gradient}: 
    preliminary neighborhood selection threshold $\in \{0.5, 2\}$, size of hidden layer $\in \{8, 32\}$,
    pruning cutoff $\in \{10^{-3}, 10^{-5}\}$

    \item IGSP \citep{wang2017permutation}: significance $\alpha \in \{10^{-2}, 10^{-3}, 10^{-4}\}$,
    CI test $\in \{\text{Gaussian, HSIC-$\gamma$}\}$

    \item PC \citep{spirtes2000causation}: 
    significance $\alpha \in \{10^{-2}, 10^{-3}, 10^{-4}\}$,
    CI test $\in \{\text{Gaussian, HSIC-$\gamma$}\}$

\end{itemize}
DAG-GNN, DCDI, DiBS, and GraN-DAG use 80\% of the available data to perform inference and compute held-out log likelihood or ELBO scores on the other 20\% of the data. 
The best hyperparameters are then selected by averaging the metric over five held-out instances of $d=30$ variables.
DiBS draws \num{10} samples from  $p(G \given D)$ using the interventional BGe score for \lineardomain and a nonlinear Gaussian interventional likelihood with MLP means for \rffdomain and \grndomain.
DiBS assumes an observation noise of \num{1}, uses a scale-free graph prior, and anneals acyclicty and relaxation parameters with rate \num{1}.
All remaining parameters are kept at the settings suggested by the authors.

For the PC algorithm and IGSP, there is no held-out score, so we compute the SID and F1 scores using the ground-truth causal graphs to select their optimal parameters. 
This would not be possible in practice and thus favors these methods.
The HSIC-$\gamma$ CI test did not scale to $d=100$ variables, so in these cases PC and IGSP always use the Gaussian CI test. 
For \grndomain $d=30$, IGSP also uses the Gaussian CI test because it OOMs at 100GB when using HSIC-$\gamma$. 
GES and GIES use the linear Gaussian BIC score function and thus do not require calibrating a sparsity parameter \citep{chickering2003optimal,hauser2012characterization}.
LiNGAM is based on independent component analysis and requires no regularization tuning either \citep{shimizu2006linear}.

\paragraph{DAG bootstrap}
To estimate edge probabilities for the non-Bayesian methods in Section \ref{ssec:benchmarking}, we use the nonparametric DAG bootstrap \citep{friedman1999data}.
We bootstrap ten datasets $D'$ from $D$ by sampling with replacement and then run each baseline individually on each bootstrapped dataset $D'$. The nonparametric probability estimate for an edge then amounts to the proportion of predicted graphs $G'$ that contain the edge.

\paragraph{Implementation}
For GES, GIES, PC, and LiNGAM, we run the original R implementations of the authors using an extended version of the software by \citet{kalainathan2020causal} (MIT Licence).
For DCDI, DAG-GNN, GraN-DAG, and DiBS, we use the Python implementations provided by the authors
\citep{brouillard2020differentiable,yu2019daggnn,lachapelle2019gradient,lorch2021dibs} (MIT License, Apache License 2.0, MIT License, MIT Licence).
For IGSP, we use the implementation provided as part of the CausalDAG package \citep{Squires18Causal} (3-Clause BSD license).

LiNGAM relies on the inversion of a covariance matrix, which frequently fails in the \grndomain domain due to the high sparsity in $D$.
Thus, to benchmark LiNGAM in \grndomain, we add small Gaussian noise to the standardized count matrix $D$.
For the IGSP and PC algorithms, the same numerical adjustment is needed to avoid crashes in the CI tests on \grndomain.
Single IGSP runs that still failed for $d=100$ were ignored when computing the metrics.
In the \grndomain results, we ignored a small number of single runs of DCDI for $d=100$ and PC for $d=30$ that failed to terminate after \num{24} hours walltime (on a GPU machine for the former).
Lastly, the CAM pruning post-processing procedure of the author's implementation of GraN-DAG \citep{lachapelle2019gradient} crashes in a few instances. We skip the post-processing step in these cases.

\section{Extended Results} \label{app:extended-results}
 
\paragraph{Compute Resources}
To carry out the experiments in this work, we trained three main \ours models and several ablations.
Each model was optimized for approximately four days using \num{8} Quadro RTX 6000 or NVIDIA GeForce RTX 3090 GPUs (\num{24} GiB memory each) and \num{128} CPUs.
To perform the benchmarking experiments, all baselines were run on four to eight CPUs each for up to \num{24} hours, depending on the method.
DCDI required one GPU to ensure a computation time of less than one day per task instance.
In all experiments, test-time inference with \ours is done on eight CPUs and no GPU.

\subsection{\ours generalization between \lineardomain and \rffdomain} \label{app:extended-results-generalization}

\input{tab_ood_linear_rff}

In this section, we provide additional out-of-distribution generalization results for \ours.
Specifically, we test the \ours model trained on the \lineardomain domain on inference from \rffdomain data, and vice versa.
This means that the \ours models not only operate under distributional shifts on the parameters of their respective data-generating processes, but also on the function classes of causal mechanisms themselves.
The models infer causal structure from data generated from function classes never seen during training.
As in all empirical analyses of Section \ref{sec:results}, the graph and noise parameters are additionally \ood, that is, the \lineardomain \ours model is tested on the \ood \rffdomain data, and vice versa.

Table \ref{tab:benchmark-transfer-linear-rff} summarizes the results.
Even under this distributional shift, the performance of both \ours models decreases reasonably and remains on par with most baselines (Table \ref{tab:benchmark-30}).
On \lineardomain data, the baselines achieve F1 scores of \num{0.15} - \num{0.54} with observational and \num{0.33} - \num{0.74} with interventional data, similar to the \rffdomain \ours model with \num{0.19} and \num{0.45}, respectively.
Conversely, on \rffdomain data, the baselines achieve F1 scores of \num{0.22} - \num{0.42} with observational and \num{0.34} - \num{0.41} with interventional data, which is also matched by the \lineardomain \ours model here with \num{0.27} and \num{0.42}, respectively.
Overall, the \lineardomain \ours model generalizes marginally better to \rffdomain data as vice versa. 
We do not report the SID here because the R code of \citet{peters2015structural} runs out of memory.

\subsection{In-Distribution Benchmarking Results for $d=30$}\label{app:extended-results-30-in-distribution}
\input{tab_benchmark_30_indist}

Table \ref{tab:benchmark-30-indist} gives the benchmarking results for in-distribution data of $d=30$ variables given the otherwise unchanged setup of Section \ref{ssec:benchmarking}.
Contrary to the \ood setting, the data is generated under homogeneous, additive noise and the parameters of their generative processes are sampled from the training domains of \ours (cf. Table \ref{tab:domain_spec_linear_rff}).
However, as throughout all experiments, the datasets and its data-generating parameters themselves are unique and have not been used by \ours during training.

Compared to the \ood setting, most baselines perform roughly the same. Since the data-generating processes are sampled from its training distribution, \ours significantly improves by moving to the easier in-distribution setting, in particular in the SCM domains, which are less noisy.
In the \grndomain domain, some baselines achieve slightly better F1 scores compared to the \ood setting, which is most likely explained by a change in the graph rather than the simulator parameter distribution, since there is no reason to believe that different generative parameters are more challenging to the baselines.

\subsection{Benchmarking Results for $d=100$}\label{app:extended-results-100}
\input{tab_benchmark_100}

Table \ref{tab:benchmark-100} shows the benchmarking results for $d=100$ variables given $n=1000$ observations and the experimental setup of Section \ref{ssec:benchmarking}.
We highlight that in this evaluation regime, \ours operates under distribution shift in terms of the causal structures, mechanisms or simulator parameters, and noise distributions, as well as the number of variables and the number of observations seen during training.

Overall, the qualitative ranking of the methods is very similar as for $d=30$.
\ours outperforms all baselines in the nonlinear \rffdomain domain, with and without access to interventional data.
Likewise, \ours is the only method to achieve nontrivial edge accuracy in terms of F1 score on the challenging \grndomain domain.
On the simpler \lineardomain domain,
there is no statistically significant difference between GES/GIES and \ours, which perform overall most favorably.

\subsection{Benchmarking Results on Real-World Proteomics Data}\label{app:extended-results-sachs}
We additionaly evaluate all of the methods on the real-world dataset by \citet{sachs2005causal}, which contains continuous measurements of $d=11$ proteins involved in human immune system cells.
Structure learning algorithms are commonly compared on this dataset, and for completeness, we report the performance of \ours and the baselines here.
However, the ground-truth network of \num{17} edges put forward by \citet{sachs2005causal} has been challenged by some experts \citep{mooij2020joint} and the assumptions of causal sufficiency and acyclicity may not be justified even though assumed by most methods, which should be kept in mind when interpreting the results.
A large part of the data are interventional, in which the measured proteins were activated or inhibited using specific reagents.
Most interventions are likely not perfect and the intervention targets may not be completely accurate \citep{mooij2020joint}.

For this experiment, we follow \citet{wang2017permutation} and \citet{brouillard2020differentiable} and discard data in which interventions were not targeted directly at one of $d=11$ measured proteins.
Given this setup, we have $n$~$=$~$5846$ data points that contain $1755$ observational and $4091$ interventional measurements, which consist of five single-protein perturbations. 
In our results, methods that only use observational data take the concatenation of all of the data without the intervention target information as input.
All baselines use the hyperparameters tuned for the nonlinear \rffdomain domain. The data is standardized to have mean \num{0} and variance \num{1}.

\input{tab_sachs}
\input{fig_sachs_preds}

Table \ref{tab:sachs} summarizes the results of all methods with respect to the reference causal graph. Figure \ref{fig:sachs-predictions} visualizes the prediction of each method.
Overall, the results are not very conclusive.
GES and GIES perform best in terms of SID, GraN-DAG is most favorable in terms of F1, and together with DCDI and \ours also in terms of SHD.
More generally, the number of predicted edges varies greatly across methods.
Almost all F1 scores fall between \num{0.25} and \num{0.30}.

\subsection{Uncertainty quantification for $d=30$}\label{app:extended-results-calibration}
\paragraph{Calibration} 
Figure \ref{fig:calibration-all} gives the calibration plots for all methods considered in the uncertainty analysis of Section~\ref{ssec:benchmarking} of the main text.
In the SCM domains, \ours closely traces the diagonal calibration line, both when given access to observational and mixed data.
Here, the nonparametric bootstraps of the PC, GIES, and IGSP algorithms as well as DiBS are similarly well-calibrated. These baselines achieve worse expected calibration error (ECE) than \ours because a significantly larger total proportion of \ours's predictions are well-calibrated (cf.\ Equation \ref{eq:calibration-ece}).
DCDI, LiNGAM, and DAG-GNN are highly overconfident, that is, they predict edges with high probability when empirically only few edges exist.

\input{fig_calibration_methods}

\paragraph{Probabilistic metrics}
Table \ref{tab:benchmark-30-prob} summarizes the probabilistic AUROC and AUPRC metrics for all methods in the experiment of Figure \ref{fig:calibration-avici}.
Explanations and interpretations for both metrics are given in Section \ref{app:metrics}.
The relative performance of the bootstrap baselines and \ours is similar to the point estimate benchmark.
Overall, \ours performs favorably across the three domains, with GES and GIES on par in \lineardomain.
However, since AUROC and AUPRC metrics evaluate the full spectrum of decision thresholds, we additionally see that \ours achieves nontrivial accuracy in \grndomain even without access to gene knockout data, indicating that \ours may provide useful information even in settings where only passive observations are available. 
This aspect is not apparent when converting the posterior probability estimates of \ours based on a single threshold and then comparing SID and F1 scores as in Table \ref{tab:benchmark-30}.

\input{tab_benchmark_30_prob.tex}

\end{document}

%% file: fig_architecture.tex
\newcommand{\figarchitectureheight}{3.3cm} 
\begin{figure}
     \centering
     \vfiguretopofpage
     \begin{subfigure}[t]{0.23\textwidth}
         \centering
         \includegraphics[height=\figarchitectureheight]{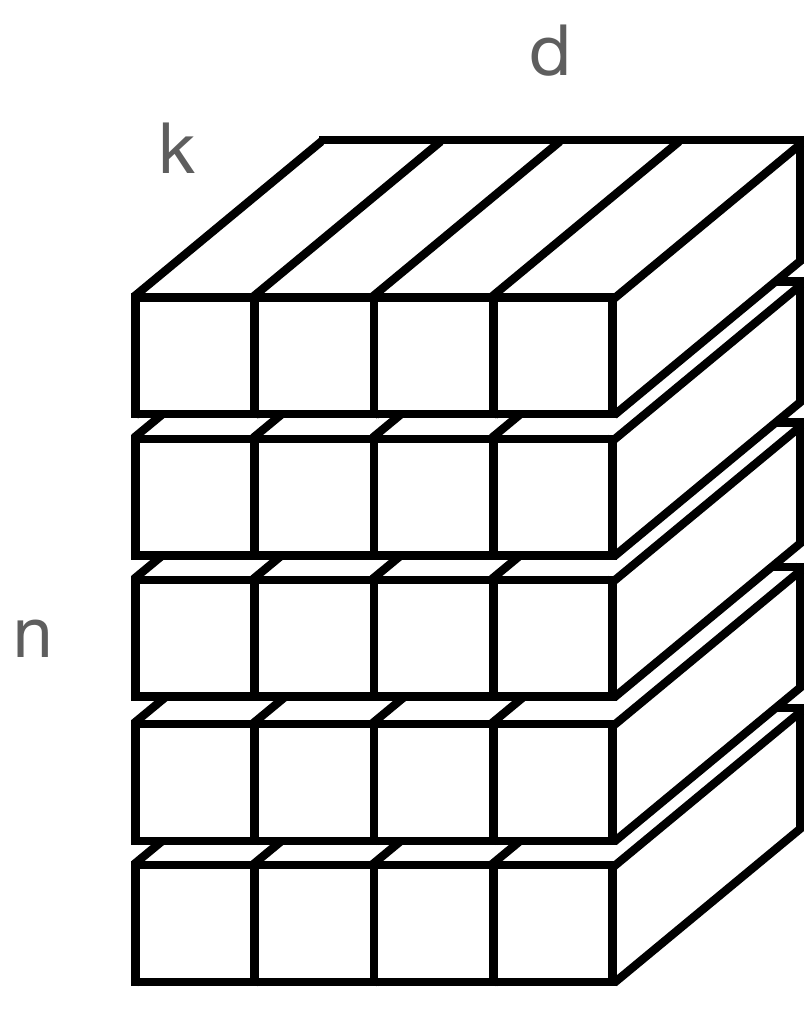}
         \caption{}
         \label{fig:architecture-a}
     \end{subfigure}
     \hfill
     \begin{subfigure}[t]{0.43\textwidth}
         \centering
         \includegraphics[height=\figarchitectureheight]{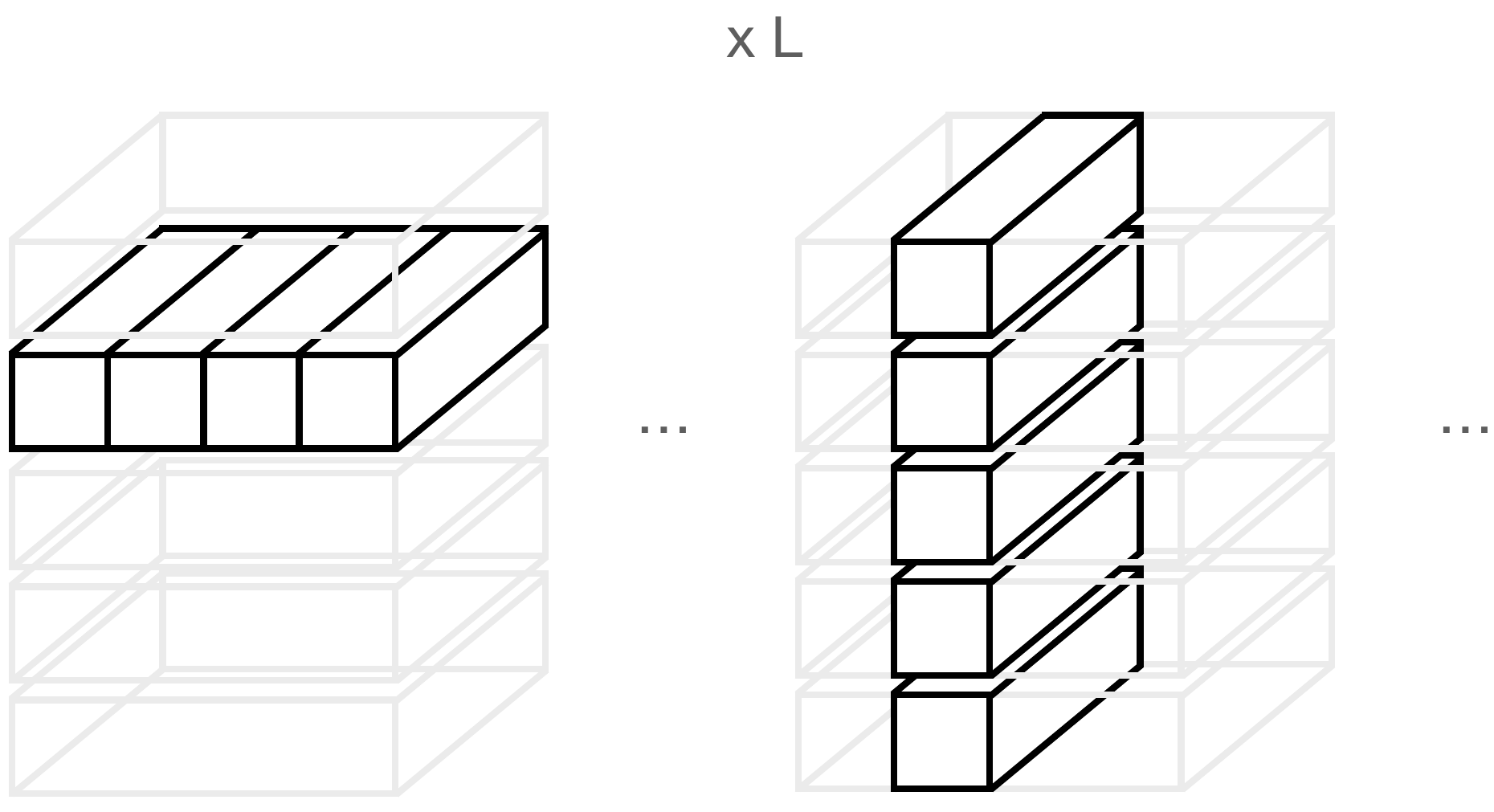}
         \caption{}
         \label{fig:architecture-b}
     \end{subfigure}
     \hfill
     \begin{subfigure}[t]{0.23\textwidth}
         \centering
         \includegraphics[height=\figarchitectureheight]{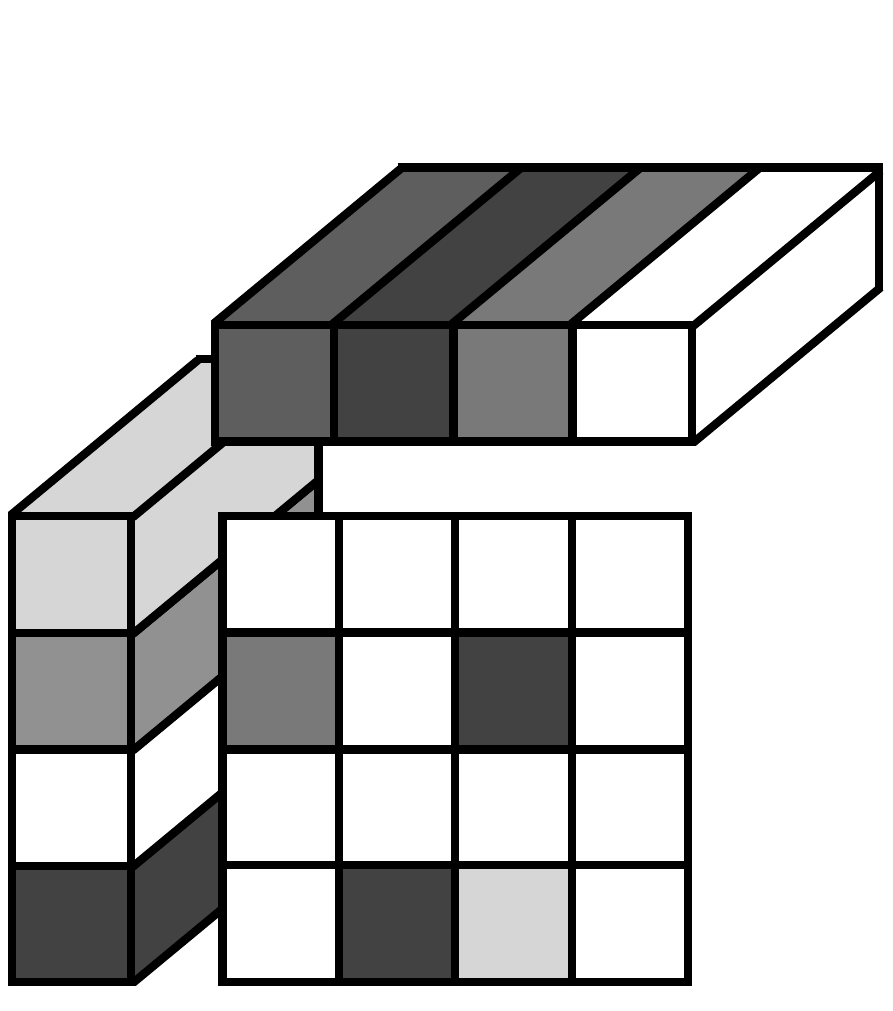}
         \caption{}
         \label{fig:architecture-c}
     \end{subfigure}
        \caption{
        \looseness-1
        {\bf Model architecture. }
        {\bf(a)}~Our model maps the input to a three dimensional tensor of shape $n \times d \times k$ and remains permutation in- and equivariant over axes $n$ and $d$, respectively. 
        {\bf(b)}~Each of the $L$ layers first self-attends over axis $d$ and then over $n$, sharing parameters across the other axis.
        {\bf(c)}~The inner product of two variables' representations
        models the probability of a direct causal effect.
        \vfigurecaptionbelow
        }
        \label{fig:architecture}
\end{figure}

%% file: fig_algo.tex
\begin{wrapfigure}[8]{r}{\algowidth} 
\vspace*{-23pt}
  \begin{minipage}[t]{\algowidth}
    \begin{algorithm}[H]
    \caption{Training the inference model~$f_\phi$}
    \begin{algorithmic}
    \STATE{Parameters: $\phi$ variational, $\lambda$ dual, $\eta$ step size}
    \WHILE{not converged}
        \FOR{$l$ \textbf{steps}}
      		\STATE{$\Delta \phi \propto \nabla_\phi \big ( 
      		     \Lcal(\phi) - \lambda \Fcal(\phi)
      		\big ) $}
    	\ENDFOR
       	\STATE{$\lambda \gets \lambda + \eta \Fcal(\phi)$}
    \ENDWHILE
    \end{algorithmic}
    \label{alg:training}
    \end{algorithm}
  \end{minipage}
\end{wrapfigure}

%% file: fig_illustration.tex
\newcommand{\figillgraph}{2.2cm} 
\newcommand{\figillfunc}{2.5cm} 
\newcommand{\figillnoise}{2.5cm} 

\newcommand{\figillspace}{-2pt} 
\begin{figure}[t]
    \centering
     \vfiguretopofpage
    \hspace*{-13pt}
    \begin{subfigure}[t]{0.47\textwidth}
        \centering
        \subfloat{
            \centering
            \includegraphics[height=\figillgraph]{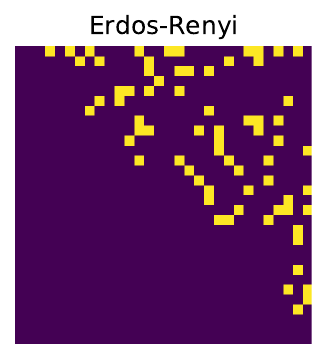}
            \hspace{\figillspace}
            \includegraphics[height=\figillgraph]{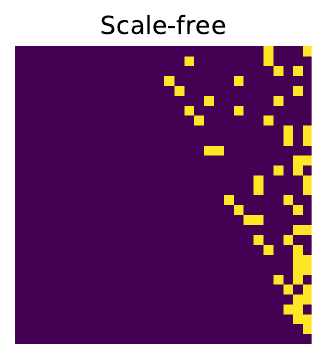}
        }
        
        \subfloat{
            \centering
            \includegraphics[height=\figillgraph]{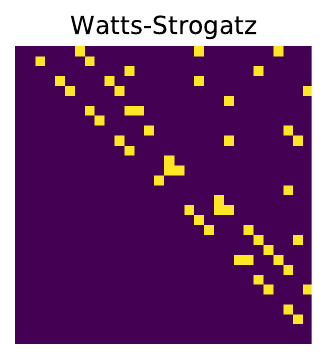}
            \hspace{\figillspace}
            \includegraphics[height=\figillgraph]{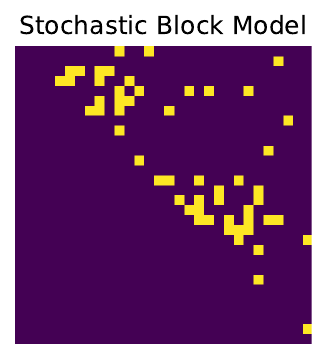}
            \hspace{\figillspace}
            \includegraphics[height=\figillgraph]{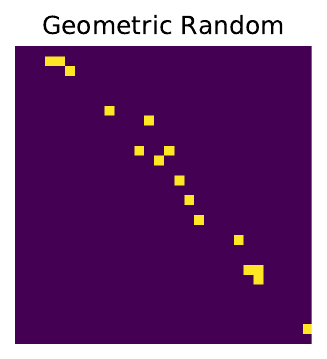}
        }
    \end{subfigure}
    \begin{subfigure}[t]{0.36\textwidth}
        \centering
        \subfloat{
            \centering
            \includegraphics[height=\figillfunc]{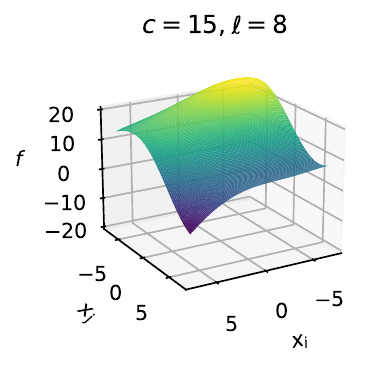}
        }
        
        \subfloat{
            \centering
            \includegraphics[height=\figillfunc]{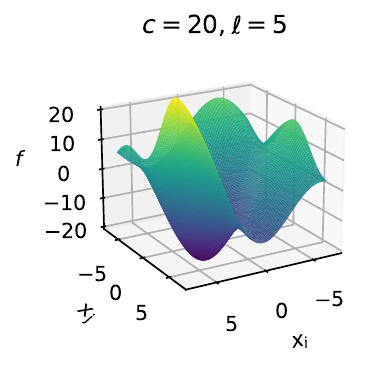}
            \hspace{\figillspace}
            \includegraphics[height=\figillfunc]{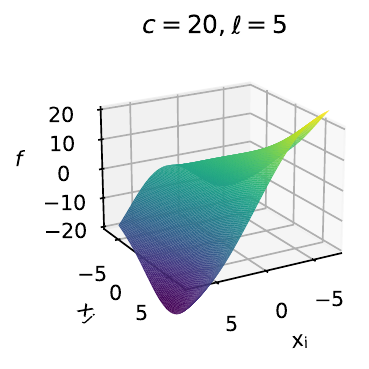}
        }
    \end{subfigure}
    \begin{subfigure}[t]{0.18\textwidth}
        \centering
        \subfloat{
            \centering
            \includegraphics[height=\figillnoise]{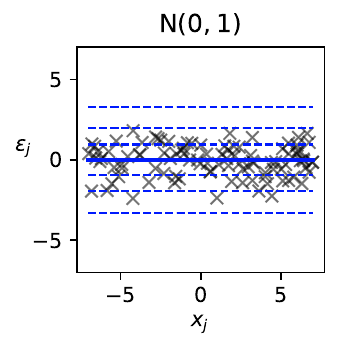}
        }
        
        \subfloat{
            \centering
            \includegraphics[height=\figillnoise]{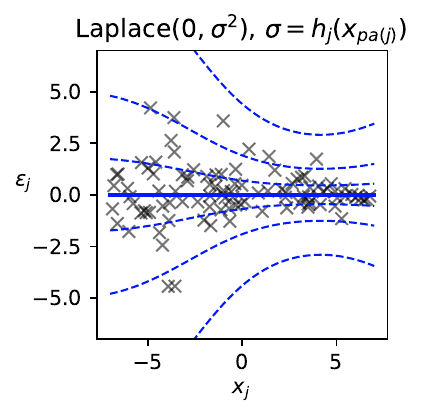}
        }
    \end{subfigure}
    \hspace*{-7pt}
    \vspace{-5pt}
    \begin{center}
        \small
        \hspace{55pt}
        \textbf{(a)}
        \hspace{160pt}
        \textbf{(b)}
        \hspace{95pt}
        \textbf{(c)}
    \end{center}
    \vspace{-5pt}
    \caption{
    \looseness - 1
    {\bf Moving out-of-distribution in the \domain{Rff} domain. }
    Randomly sampled data-generating components of the nonlinear SCM domain during training $p(D)$ (top) and \ood evaluation $\tilde{p}(D)$ (bottom).
    For visualization, 
    the adjacency matrices {\bf(a)} are topologically sorted,
    the causal mechanisms {\bf(b)} have two parents, where $c$ and $\ell$ are output and length scales of the underlying GP,
    and the noise {\bf(c)} is shown as a function of one parent, where dashed lines indicate \num{0.66}, \num{0.95}, and \num{0.999} coverage.
    \vfigurecaptionbelow
    }
    \label{fig:domain-illustration}
\end{figure}

%% file: fig_results_generalization.tex
\newcommand{\figheightradar}{4.7cm} 
\newcommand{\figheightgeneral}{4.1cm} 

\newcommand{\figradarhspace}{-5pt} 
\newcommand{\figgeneralhspace}{-17pt} 
\begin{figure}[t]
    \centering
     \vfiguretopofpage
    \begin{subfigure}[t]{0.32\textwidth}
        \centering
        \subfloat{
            \centering
            \hspace*{\figradarhspace}\includegraphics[height=\figheightradar]{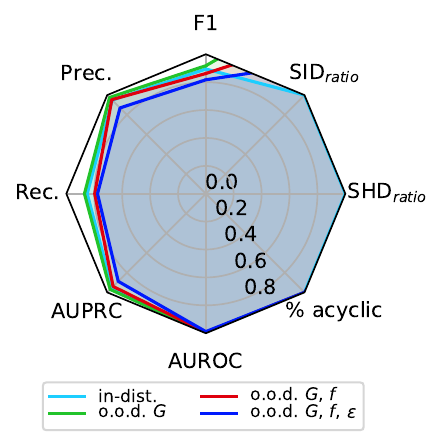}
        }
        
        \subfloat{
            \centering
            \hspace*{\figgeneralhspace}\includegraphics[height=\figheightgeneral]{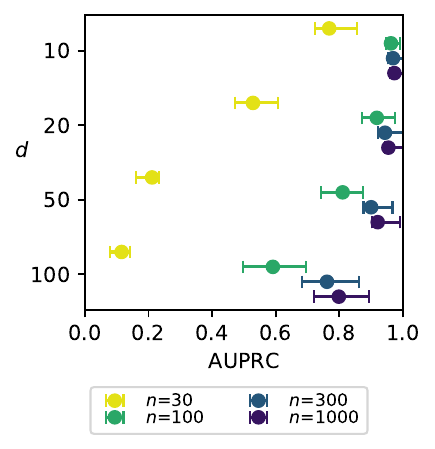}
        }
        \addtocounter{subfigure}{-2}
        \caption{\lineardomain}
    \end{subfigure}
    \hfill
    \begin{subfigure}[t]{0.32\textwidth}
        \centering
        \subfloat{
            \centering
            \hspace*{\figradarhspace}\includegraphics[height=\figheightradar]{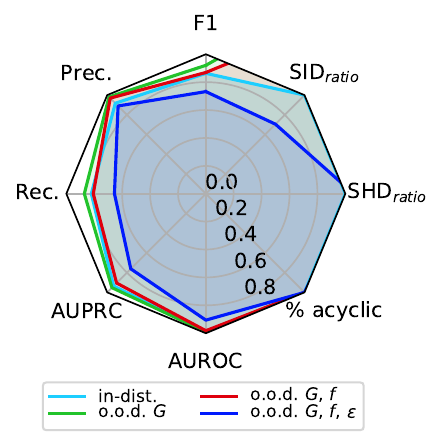}
        }
        
        \subfloat{
            \centering
            \hspace*{\figgeneralhspace}\includegraphics[height=\figheightgeneral]{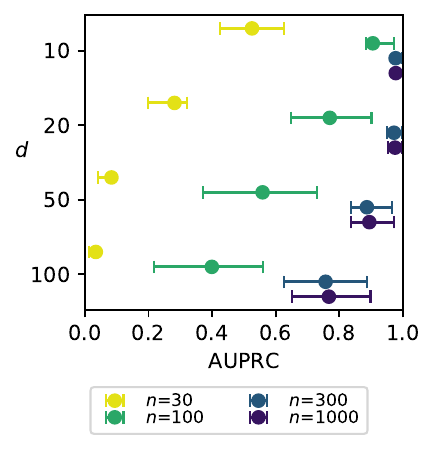}
        }
        \addtocounter{subfigure}{-2}
        \caption{\rffdomain}
    \end{subfigure}
    \hfill
    \begin{subfigure}[t]{0.32\textwidth}
        \centering
        \subfloat{
            \centering
            \hspace*{\figradarhspace}\includegraphics[height=\figheightradar]{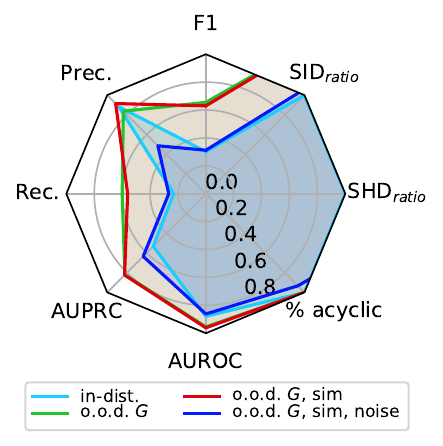}
        }
        
        \subfloat{
            \centering
            \hspace*{\figgeneralhspace}\includegraphics[height=\figheightgeneral]{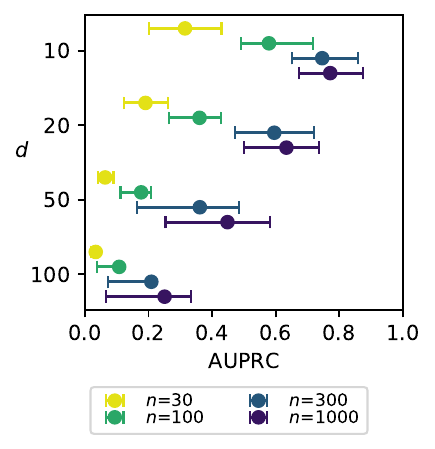}
        }
        \addtocounter{subfigure}{-2}
        \caption{\grndomain}
    \end{subfigure}
    \caption{
    \looseness - 1
    {\bf Generalization properties of the inference model $f_\phi$. }
    Top row plots show performance metrics of \ours under increasing distributional shift given $n$~$=$~$1000$ observations for $d$~$=$~$30$ variables.
    $\sidratio$ is defined as $\text{SID}_\text{in-dist.}/\text{SID}$ and analogously for $\text{SHD}$. Thus, higher is better for all metrics.
    Bottom row shows the in-distribution AUPRC for various $d$ as we vary the number of observations provided to \ours.
    The datasets contain interventional data (cf.~Section \ref{ssec:domains}).
    All values are the mean over fifteen random task instances.
    Error bars indicate the interquartile range.
    \vfigurecaptionbelow
    }
    \label{fig:generalization}
\end{figure}

%% file: tab_benchmark_30.tex
\begin{table}[t]
\vfiguretopofpage
\vspace*{5pt}
\caption{\looseness - 1 
{\bf Benchmarking results ($d=30$ variables). }
Mean SID ($\downarrow$) and F1 score ($\uparrow$) with standard error of all methods on \num{30} random task instances. 
Methods in the top section use only observational data, in the bottom section both observational and interventional data.
We highlight the best result of each section and those within its \num{95}\% confidence interval according to an unequal variances $t$-test.
}\label{tab:benchmark-30}
\vspace{7pt}
\centering
\begin{adjustbox}{max width=0.90\linewidth}
\begin{threeparttable}
\begin{tabular}{lcccccc}
\toprule
 &           \multicolumn{2}{c}{\domain{Linear}}            &            \multicolumn{2}{c}{\domain{Rff}}           &        \multicolumn{2}{c}{\domain{Grn}}       \\
\cmidrule{2-7}
Algorithm &              SID &               F1  &              SID &               F1 &              SID &               F1\\
\midrule
\ges               &  \highlight{215.6 \err{35.0}} &              0.548 \err{0.03} &  \highlight{346.3 \err{44.4}} &              0.285 \err{0.03}  &  \highlight{573.6 \err{29.2}} &  \highlight{0.058 \err{0.01}} \\  
\lingam            &              413.4 \err{48.4} &              0.369 \err{0.04} &              410.3 \err{47.6} &              0.238 \err{0.02}  &  \highlight{617.5 \err{31.7}} &  \highlight{0.044 \err{0.01}} \\  
\pc                &              400.5 \err{53.7} &              0.338 \err{0.03} &  \highlight{370.1 \err{51.2}} &              0.421 \err{0.03}  &  \highlight{594.0 \err{30.0}} &  \highlight{0.061 \err{0.01}} \\  
\daggnn            &              474.5 \err{50.8} &              0.154 \err{0.01} &              425.3 \err{50.2} &              0.221 \err{0.03}  &  \highlight{588.7 \err{36.6}} &  \highlight{0.078 \err{0.02}} \\  
\grandag           &              466.0 \err{54.3} &              0.200 \err{0.03} & \highlight{328.6 \err{48.4}} &  \highlight{0.476 \err{0.05}} & \highlight{582.4 \err{33.4}} & \highlight{0.073 \err{0.02}}\\
\oursobserv  &  \highlight{145.6 \err{21.5}} &  \highlight{0.672 \err{0.04}} &  \highlight{255.1 \err{48.2}} &  \highlight{0.618 \err{0.06}}  &  \highlight{641.7 \err{34.7}} &              0.000 \err{0.00} \\  
\midrule
\gies              &  \highlight{120.8 \err{26.2}} &              0.736 \err{0.03} &  \highlight{304.8 \err{44.0}} &              0.338 \err{0.04}  &              545.5 \err{26.9} &              0.092 \err{0.01} \\  
\igsp              &              244.0 \err{34.4} &              0.559 \err{0.02} &              374.1 \err{45.0} &              0.407 \err{0.04}  &              597.4 \err{31.7} &              0.057 \err{0.01} \\  
\dcdi              &              383.5 \err{45.1} &              0.327 \err{0.03} &  \highlight{282.8 \err{46.3}} &              0.409 \err{0.04}  &              590.9 \err{30.6} &              0.075 \err{0.02} \\  
\oursinterv &  \highlight{110.9 \err{19.3}} &  \highlight{0.819 \err{0.02}} &  \highlight{192.7 \err{44.8}} &  \highlight{0.707 \err{0.06}}  &  \highlight{416.9 \err{47.1}} &  \highlight{0.338 \err{0.06}} \\  
\bottomrule
\end{tabular}
\end{threeparttable}
\end{adjustbox}
\vfigurecaptionbelow
\vspace*{0pt}
\end{table}

%% file: fig_calibration_combined.tex
\newcommand{\figheightcalibration}{5.2cm} 
\newcommand{\figcalibrationhspace}{10pt} 
\newcommand{\calibrationfigurerelwidth}{0.46} 
\newcommand{\calibrationtablerelwidth}{0.49} 

\addtocounter{figure}{-1}    

\begin{figure}[t]
    \centering
    \vfiguretopofpage
    \vspace*{0pt}
    \begin{subfigure}[t]{\calibrationfigurerelwidth\textwidth}
        \hspace*{\figcalibrationhspace}\includegraphics[height=\figheightcalibration]{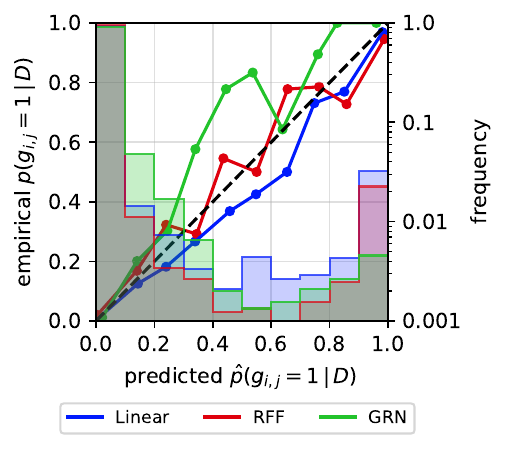}
    \label{fig:calibration-avici-a}
    \vspace*{-4pt}
    \caption{}
    \end{subfigure}
    \hfill
    %
    \begin{adjustbox}{max width=\calibrationtablerelwidth\linewidth}
    \begin{threeparttable}
    \vspace*{-167pt}
    \input{tab_calibration}

    \begin{center}
        \vspace*{7pt}
        {\bf (b)}
        \vspace*{2pt}
    \end{center}
    \end{threeparttable}
    \end{adjustbox}
    \vspace*{-4pt}
    \caption{
    \looseness-1
    {\bf Uncertainty calibration ($d=30$ variables). }
    As previously, datasets are held-out and \ood in terms of graph, parameters, and noise.
    {\bf (a)}
    Calibration plots for \ours aggregating the predictions for ten test cases of each domain. 
    The histograms on the right $y$-axis show the frequency of predictions at each confidence level.
    {\bf (b)}
    ECE ($\downarrow$) with standard error averaged over ten test cases.
    Methods in the top (bottom) section use observational (and interventional) data.
    We highlight the best result and those within its \num{95}\% confidence interval according to an unequal variances $t$-test.
    \vfigurecaptionbelow
    }
    \label{fig:calibration-avici}
\end{figure}

%% file: tab_calibration.tex
\begin{tabular}{lccc}
\toprule
& \domain{Linear} & {\domain{Rff}} & \domain{Grn} \\ 
\midrule
\ges${}^*$    &              0.031 \err{0.00} &              0.068 \err{0.02} &              0.092 \err{0.01} \\
\lingam${}^*$ &              0.066 \err{0.02} &              0.054 \err{0.01} &              0.053 \err{0.01} \\
\pc${}^*$     &              0.036 \err{0.00} &  \highlight{0.033 \err{0.01}} &              0.065 \err{0.01} \\
\daggnn${}^*$ &              0.078 \err{0.01} &              0.063 \err{0.01} &              0.063 \err{0.01} \\
\grandag${}^*$&              0.046 \err{0.01} &  \highlight{0.042 \err{0.01}} &              0.199 \err{0.05}\\
\oursobserv      &  \highlight{0.013 \err{0.00}} &  \highlight{0.024 \err{0.01}} &  \highlight{0.018 \err{0.00}} \\
\midrule
\gies${}^*$   &              0.027 \err{0.00} &              0.074 \err{0.02} &              0.094 \err{0.01} \\
\igsp${}^*$   &              0.042 \err{0.01} &              0.083 \err{0.01} &              0.077 \err{0.01} \\
\dcdi${}^*$   &              0.068 \err{0.01} &              0.087 \err{0.02} &              0.170 \err{0.03} \\
\dibs             &              0.056 \err{0.02} &  \highlight{0.035 \err{0.01}} &              0.093 \err{0.01} \\
\oursinterv      &  \highlight{0.011 \err{0.00}} &  \highlight{0.022 \err{0.01}} &  \highlight{0.024 \err{0.01}} \\
\bottomrule
\end{tabular}
\begin{tablenotes}
\footnotesize
\item[$*$] Nonparametric DAG bootstrap \citep{friedman1999data}
\end{tablenotes}

%% file: tab_ablation.tex
\begin{table}[t]
\vfiguretopofpage
%
\caption{\looseness - 1 
{\bf Ablations of the architecture of $f_\phi$. }
Models are evaluated 
on \num{100} interventional datasets of $d=30$ variables in the \rffdomain domain.
Top row ($\star$) corresponds to the model used in the main experiments.
In (a), we vary the number of blocks $L$; 
in (b), the axes over which attention is performed; 
in (c), the generative model of the variational parameters;
in (d), the number of update steps of $\phi$.
We again highlight the best result and those within its \num{95}\% $t$-test confidence interval.
}\label{tab:ablation}
\vspace{7pt}
\centering
\begin{adjustbox}{max width=0.97\linewidth}
\begin{threeparttable}
\begin{tabular}{cccccc cccc}
\toprule
&&&&&&\multicolumn{2}{c}{\rffdomain (in-dist.)} & \multicolumn{2}{c}{\rffdomain (\ood)}\\
\cmidrule{7-10}
& $L$ & ax.~$d$ & ax.~$n$ & $\theta$ model & steps & SID  & AUPRC &  SID & AUPRC \\
\midrule
($\star$) &   8 & \checkmark & \checkmark & Eq. (\ref{eq:dibs}) & 300k  &    \highlight{65.2 \err{8.4}} &  \highlight{0.972 \err{0.00}}  &  \highlight{221.5 \err{24.7}} &  \highlight{0.650 \err{0.03}}  \\
\midrule 
\multirow{3}{*}{(a)}
&   1 &            &            &                     & &              267.2 \err{22.0} &              0.635 \err{0.01}  &              394.2 \err{28.4} &              0.242 \err{0.02}  \\
&   2 &            &            &                     & &              195.9 \err{18.5} &              0.825 \err{0.01}  &              343.1 \err{27.1} &              0.400 \err{0.03}  \\
&   4 &            &            &                     & &              116.6 \err{13.1} &              0.937 \err{0.01}  &  \highlight{264.0 \err{24.8}} &              0.566 \err{0.03}  \\
\midrule 
\multirow{2}{*}{(b)} 
&    & \checkmark &            &                     & &              351.5 \err{27.9} &              0.552 \err{0.01}  &              414.2 \err{29.5} &              0.209 \err{0.02}  \\
&     &            & \checkmark &                     & &              416.8 \err{29.6} &              0.256 \err{0.01}  &              390.2 \err{27.6} &              0.078 \err{0.01}  \\
\midrule 
\multirow{1}{*}{(c)} 
&     &            &            & \hspace*{-3pt}{\small\citep{santoro2017simple}}\hspace*{-3pt} &   & \highlight{72.4 \err{9.2}} &  \highlight{0.971 \err{0.00}}  &  \highlight{225.7 \err{25.2}} &  \highlight{0.634 \err{0.03}}  \\
\midrule 
\multirow{1}{*}{(d)} 
&     &            &            & & 100k &               96.9 \err{11.6} &              0.955 \err{0.00}  &  \highlight{259.3 \err{26.6}} &  \highlight{0.589 \err{0.04}}  \\
\bottomrule
\end{tabular}
\end{threeparttable}
\end{adjustbox}
\vfigurecaptionbelow
\end{table}





%% file: tab_domain_spec_rff.tex
\begin{table}[h]
    \centering
    \caption{\looseness -1 Specification of the training and out-of-distribution data-generating processes $p(D)$ and $\tilde{p}(D)$ for the \domain{Linear} and \domain{Rff} domain. All specifications except the mechanism function type are the same for the two domains.  For each random task instance, the parameter configurations are sampled uniformly randomly from all possible combinations of the sets of options. The graph model classes are sampled in equal proportions out-of-distribution. Empty fields indicate that the component is not part of the distribution.}\label{tab:domain_spec_linear_rff}
    \vspace{15pt}
\begin{adjustbox}{max width=\linewidth}
\begin{threeparttable}
\begin{tabular}{lllll}
\toprule
 & \multicolumn{2}{c}{\domain{In-distribution $p(D)$}}
 & \multicolumn{2}{c}{\domain{Out-of-distribution $\tilde{p}(D)$}}
\\
\cmidrule{2-5}
\textbf{Graph} & & & &  \\
\midrule
\erdosrenyi & expected edges/node & $\in \{1, 2, 3\}$ &  & \medskip\\
Scale-free (in-degree) & edges/node & $\in \{1, 2, 3\}$ &  & \\
 & attach.\ power $\alpha$  & $\in \{1.0\}$ &  & \medskip\\
Scale-free (out-degree) & edges/node & $\in \{1, 2, 3\}$ & edges/node & $\in \{2\}$\\
 & attach.\ power $\alpha$ & $\in \{1.0\}$ & attach.\ power $\alpha$  & $\in \{0.5, 1.5\}$ \medskip\\
Watts-Strogatz &&& lattice dim.\ $k$ & $\in \{2, 3\}$ \\
&&& rewire prob.\ & $\in \{0.3\}$ \medskip\\
Stochastic Block Model &&& expected edges/node & $\in \{2\}$ \\
&&& blocks & $\in \{5, 10\}$ \\
&&& damp.\ inter-block prob. & $\in \{0.1\}$ \medskip\\
Geometric Random Graphs &&& radius & $\in \{0.1\}$ \\
\midrule
\\
\textbf{Mechanism} & & & &  \\
\midrule
Linear function${}^{\text{(a)}}$ & weights $\wb$ & $\sim \unifpm(1, 3)$ & weights $\wb$ & $\sim \unifpm(0.5, 2)$ \\
&&&& $\sim \unifpm(2, 4)$ \\
& bias $b$ & $\sim \unif(-3, 3)$ & bias $b$ & $\sim \unif(-3, 3)$ \\
\midrule
Random Fourier function${}^{\text{(b)}}$ & SE length scale $\ell$ & $\sim \unif(7, 10)$ & SE length scale $\ell$ & $\sim \unif(5, 8)$ \\
&&&& $\sim \unif(8, 12)$ \\
& SE output scale $c$ & $\sim \unif(10, 20)$ & SE output scale $c$ & $\sim \unif(8, 12)$ \\
&&&& $\sim \unif(18, 22)$ \\
& bias $b$ & $\sim \unif(-3, 3)$ & bias  $b$ & $\sim \unif(-3, 3)$ \\
\midrule
\\
\textbf{Noise} (indiv.\ per variable) & & & &  \\
\midrule
$\Ncal(0, \sigma^2)$& $\sigma$ & $\sim \unif(0.2, 2)$ && \\
$\text{Laplace}(0, \sigma^2)$ &&&$\sigma^2(\xb_{\text{pa}_j}) $ (heterosced.) & $ \sim p(h_\text{rff})$\\
$\text{Cauchy}(0, \sigma^2)$ &&&$\sigma^2(\xb_{\text{pa}_j}) $ (heterosced.) & $ \sim p(h_\text{rff})$\\
\midrule
\\
\textbf{Interventions} & & & &  \\
\midrule
Target nodes & random 50\% of nodes && random 50\% of nodes&\\
Intervention values & $x_j$ & $\sim \unifpm(1, 3)$& $x_j$ & $\sim \unifpm(1, 5)$\\
\bottomrule
\end{tabular}
\vspace*{5pt}
\begin{tablenotes}
\footnotesize
\item[$\text{(a)}$] Only \domain{Linear} domain
\item[$\text{(b)}$] Only \domain{Rff} domain
\vspace*{5pt}
\item[] \hspace*{-10pt}Aliases:
\begin{itemize}[leftmargin=3pt]
	\item $\unifpm(a, b)$: uniform mixture of $\unif(a, b)$ and $\unif(-b, -a)$
	\item $p(h_\text{rff})$: distribution over heteroscedastic noise scale functions, induced by the squash function $h_\text{rff}(\xb) = \log(1+\exp(g_\text{rff}(\xb))$ and random Fourier feature functions $g_\text{rff}(\xb)$ with SE length scale $\ell=10$ and output scale $c=2$ (cf.\ \domain{Rff} domain)
\end{itemize}
\end{tablenotes}
\end{threeparttable}
\end{adjustbox}
\vfigurecaptionbelow
\vspace{5pt}
\end{table}

%% file: tab_domain_spec_gene.tex
\begin{table}[p]
    \centering
    \caption{\looseness -1 Specification of the training and out-of-distribution data-generating processes $p(D)$ and $\tilde{p}(D)$ for the \domain{Gene} domain.  For each random task instance, the parameter configurations are sampled uniformly randomly from all possible combinations of the sets of options. The graph model classes are sampled in equal proportions out-of-distribution. Empty fields indicate that the component is not part of the distribution.}\label{tab:domain_spec_gene}
    \vspace{15pt}
\begin{adjustbox}{max width=\linewidth}
\begin{threeparttable}
\begin{tabular}{lllll}
\toprule
 & \multicolumn{2}{c}{\domain{In-distribution $p(D)$}}
 & \multicolumn{2}{c}{\domain{Out-of-distribution $\tilde{p}(D)$}}
\\
\cmidrule{2-5}
\textbf{Graph} & & & &  \\
\midrule
\erdosrenyi & expected edges/node & $\in \{1, 2, 3\}$ &  & \\
Scale-free (out-degree) & edges/node & $\in \{1, 2, 3\}$ & & \\
 & attach.\ power $\alpha$  & $\in \{0.5, 0.8, 1.0, 1.2, 1.5\}$ & &  \\
\ecoli subgraph &&& top-$p$ perc. modular & $\in \{0.2\}$ \\
\citep{marbach2009generating}&&&& \\
\yeast subgraph &&& top-$p$ perc. modular & $\in \{0.2\}$ \\
\citep{marbach2009generating}&&&& \\
\midrule
\\
\textbf{Mechanism} & & & &  \\
\midrule
GRN simulator & no.\ cell types & $\in \{5 \}$ & no.\ cell types & $\in \{ 10 \}$ \\
 \citep{dibaeinia2020sergio} & decay rates $\lambda$ & $\in \{0.7, 0.8, 0.9 \}$&  decay rates $\lambda$ & $\in \{0.5, 1.5\}$\\
& system noise scale $\xi$ & $\in \{0.9, 1.0, 1.1 \}$&  system noise scale $\xi$ & $\in \{0.5, 1.5\}$\\
& Hill function coeff.\ $\gamma$ & $\in \{1.9, 2.0, 2.1 \}$&  Hill function coeff.\ $\gamma$ & $\in \{1.5, 2.5 \}$\\
& MR prod.\ rate $b$ & $\sim \unif(1, 3)$ &  MR prod.\ rate $b$ & $\sim \unif(0.5, 2)$  \\
& &&& $\sim \unif(2, 4)$  \\
& interactions $k$ & $\sim \unif(1, 5)$ &  interactions $k$ & $\sim \unif(1, 3)$  \\
& &&& $\sim \unif(3, 7)$  \\
& $\text{signs}(k)$ per node & $\sim \text{Bern}(p)$ where $p$ &   $\text{signs}(k)$ per node & from \ecoli  \\
&&~~~$\sim \text{Beta}(1, 1)$ & & or $\sim \text{Bern}(p)$ \\
&&~~~$\sim \text{Beta}(0.5, 0.5)$ &&(cf.\ Sec.~\ref{app:subgraph}) \\
\midrule
\\
\textbf{Measurement Noise} & & & &  \\
\midrule
Platform${}^{\dagger}$ & 10X chromium &$p_{\text{outlier}} \in \{0.01\}$&\\
&&$\mu_{\text{outlier}} \in \{3.0, 5.0\}$&&\\
&&$\sigma_{\text{outlier}} \in \{1.0\}$&&\\
&&$\mu_{\text{lib}} \in \{4.5, 6.0\}$&&\\
&&$\sigma_{\text{lib}} \in \{0.3, 0.4, 0.7 \}$&&\\
&&$\delta \in \{ 45, 74, 82\}$&&\\
&&$\tau \in \{8.0\}$&&\medskip\\
&&&Illumina HiSeq2000&$p_{\text{outlier}} \in \{ 0.01 \}$\\
&&&&$\mu_{\text{outlier}} \in \{ 0.8\}$\\
&&&&$\sigma_{\text{outlier}} \in \{1.0 \}$\\
&&&&$\mu_{\text{lib}} \in \{ 7.0\}$\\
&&&&$\sigma_{\text{lib}} \in \{0.4\}$\\
&&&&$\delta \in \{ 80\}$\\
&&&&$\tau \in \{8.0 \}$\medskip\\
&&&Drop-seq&$p_{\text{outlier}} \in \{ 0.01\}$\\
&&&&$\mu_{\text{outlier}} \in \{3.0 \}$\\
&&&&$\sigma_{\text{outlier}} \in \{1.0 \}$\\
&&&&$\mu_{\text{lib}} \in \{ 4.4\}$\\
&&&&$\sigma_{\text{lib}} \in \{0.8\}$\\
&&&&$\delta \in \{ 85\}$\\
&&&&$\tau \in \{8.0 \}$\medskip\\
&&&Smart-seq& $p_{\text{outlier}} \in \{0.01 \}$\\
&&&&$\mu_{\text{outlier}} \in \{ 4.5\}$\\
&&&&$\sigma_{\text{outlier}} \in \{ 1.0\}$\\
&&&&$\mu_{\text{lib}} \in \{10.8 \}$\\
&&&&$\sigma_{\text{lib}} \in \{0.55\}$\\
&&&&$\delta \in \{92 \}$\\
&&&&$\tau \in \{2.0 \}$\\
\textbf{Interventions} & & & &  \\
\midrule
Target nodes & all nodes &&  all nodes &\\
Intervention type & gene knockout & & gene knockout & \\
\bottomrule
\end{tabular}
\vspace*{0pt}
\begin{tablenotes}
\footnotesize
\vspace*{5pt}
\item[$\dagger$] Noise specifications were collected from calibrations performed by \citet{dibaeinia2020sergio} on real datasets generated by the different \scrna platforms.
\end{tablenotes}
\end{threeparttable}
\end{adjustbox}
\vfigurecaptionbelow
\vspace{5pt}
\end{table}

%% file: tab_ood_linear_rff.tex
\begin{table}[t]
\vspace*{5pt}
\caption{\looseness - 1 
{\bf Generalizing from \lineardomain to \rffdomain and vice versa ($d$~$=$~$30$ variables). }
Mean SHD ($\downarrow$), F1 score ($\uparrow$), AUROC ($\uparrow$), and  AUPRC ($\uparrow$) with standard error on \num{30} random task instances. 
The domain in parentheses indicates the training domain, and the header indicates the test domain. 
We highlight the rows in which models were evaluated on the same function class as during training, though as in all benchmarking experiments, all test datasets $D$ are sampled from the \ood data-generating distributions. 
The metrics of the baselines corresponding to these experiments are given in Table~\ref{tab:benchmark-30}.
}\label{tab:benchmark-transfer-linear-rff}
\vspace{7pt}
\centering
\begin{adjustbox}{max width=\linewidth}
\begin{threeparttable}
\begin{tabular}{lccccc}
\toprule
& \multicolumn{4}{c}{\domain{Linear}}  \\
\cmidrule{2-5}
 & SHD & F1 & AUROC & AUPRC \\
\midrule
\textbf{\ours} (trained on \lineardomain) ${}^\dagger$  & \highlight{18.9 \err{2.1}}  & \highlight{0.672 \err{0.04}} &  \highlight{0.978 \err{0.00}} &  \highlight{0.790 \err{0.03}} \\ 
\textbf{\ours} (trained on \rffdomain) ${}^\dagger$  &  93.4 \err{20.1} &  0.191 \err{0.03} &  0.686 \err{0.03} &  0.179 \err{0.02}  \\ 
\midrule
\textbf{\ours} (trained on \lineardomain)  & \highlight{13.2 \err{1.8}}  &  \highlight{0.819 \err{0.02}} &  \highlight{0.988 \err{0.00}} &  \highlight{0.892 \err{0.02}}\\ 
\textbf{\ours} (trained on \rffdomain)  &  63.6 \err{14.5} &  0.452 \err{0.05} &  0.802 \err{0.03} &  0.469 \err{0.06} \\ 
\midrule
& \multicolumn{4}{c}{\domain{Rff}} \\
\cmidrule{2-5}
 & SHD & F1 & AUROC & AUPRC   \\
\midrule
\textbf{\ours} (trained on \lineardomain) ${}^\dagger$ &              36.4 \err{3.2} &              0.273 \err{0.04} &  0.784 \err{0.03} &              0.385 \err{0.04}  \\ 
\textbf{\ours} (trained on \rffdomain) ${}^\dagger$ &  \highlight{21.6 \err{3.5}} &  \highlight{0.618 \err{0.06}} &  \highlight{0.854 \err{0.03}} &  \highlight{0.659 \err{0.06}}  \\ 
\midrule
\textbf{\ours} (trained on \lineardomain)  &              34.3 \err{3.6} &              0.420 \err{0.05} & 0.811 \err{0.03} &              0.495 \err{0.05}  \\ 
\textbf{\ours} (trained on \rffdomain) &  \highlight{18.0 \err{3.6}} &  \highlight{0.707 \err{0.06}} &  \highlight{0.888 \err{0.03}} &  \highlight{0.739 \err{0.06}}  \\ 
\bottomrule
\end{tabular}
\begin{tablenotes}
\footnotesize
\item[$\dagger$] Only using observational data.
\end{tablenotes}
\end{threeparttable}
\end{adjustbox}
\vspace*{0pt}
\end{table}

%% file: tab_benchmark_30_indist.tex
\begin{table}[t]
\vspace*{5pt}
\caption{\looseness - 1 
{\bf In-distribution benchmarking results ($d=30$ variables). }
Mean SID ($\downarrow$) and F1 score ($\uparrow$) with standard error of all methods on \num{30} random task instances. 
Methods in the top section use only observational data, in the bottom section both observational and interventional data.
The best results of each section are highlighted together with those inside its \num{95}\% confidence interval according to an unequal variances $t$-test.
}\label{tab:benchmark-30-indist}
\vspace{7pt}
\centering
\begin{adjustbox}{max width=0.95\linewidth}
\begin{threeparttable}
\begin{tabular}{lcccccc}
\toprule
 &           \multicolumn{2}{c}{\domain{Linear}}            &            \multicolumn{2}{c}{\domain{Rff}}           &        \multicolumn{2}{c}{\domain{Grn}}       \\
\cmidrule{2-7}
Algorithm &              SID &               F1  &              SID &               F1 &              SID &               F1\\
\midrule
\ges         &  \highlight{217.5 \err{38.3}} &              0.643 \err{0.04} &              296.9 \err{42.6} &              0.428 \err{0.03} &  \highlight{535.3 \err{44.8}} &  \highlight{0.147 \err{0.01}} \\ 
\lingam      &              500.8 \err{43.5} &              0.161 \err{0.03} &              406.0 \err{42.3} &              0.237 \err{0.02} &  \highlight{590.7 \err{41.8}} &              0.110 \err{0.01} \\ 
\pc          &              383.5 \err{57.3} &              0.386 \err{0.04} &              386.7 \err{55.8} &              0.431 \err{0.03} &  \highlight{579.7 \err{41.9}} &  \highlight{0.128 \err{0.01}} \\ 
\daggnn      &              533.2 \err{47.1} &              0.127 \err{0.02} &              386.1 \err{38.7} &              0.252 \err{0.02} &  \highlight{576.3 \err{43.2}} &  \highlight{0.159 \err{0.02}} \\ 
\grandag     &              417.2 \err{54.1} &              0.249 \err{0.02} &              312.0 \err{38.9} &              0.477 \err{0.02} &  \highlight{572.6 \err{44.9}} &              0.079 \err{0.01} \\ 
\oursobserv &  \highlight{178.2 \err{36.9}} &  \highlight{0.828 \err{0.03}} &  \highlight{137.8 \err{29.1}} &  \highlight{0.838 \err{0.02}} &  \highlight{607.4 \err{46.2}} &              0.064 \err{0.02} \\ 
\midrule
\gies        &   \highlight{16.5 \err{10.0}} &  \highlight{0.942 \err{0.01}} &              290.4 \err{40.8} &              0.389 \err{0.02} &  \highlight{520.4 \err{45.9}} &              0.162 \err{0.02} \\ 
\igsp        &              277.6 \err{36.9} &              0.512 \err{0.04} &              386.5 \err{44.7} &              0.391 \err{0.03} &  \highlight{591.9 \err{46.1}} &              0.113 \err{0.01} \\ 
\dcdi        &              242.7 \err{36.7} &              0.559 \err{0.03} &              152.4 \err{21.6} &              0.555 \err{0.02} &              624.4 \err{38.7} &              0.080 \err{0.01} \\ 
\oursinterv &               72.4 \err{20.7} &  \highlight{0.948 \err{0.01}} &   \highlight{67.4 \err{16.8}} &  \highlight{0.927 \err{0.01}} &  \highlight{466.5 \err{49.7}} &  \highlight{0.316 \err{0.05}} \\ 
\bottomrule
\end{tabular}
\end{threeparttable}
\end{adjustbox}
\vspace*{0pt}
\end{table}

%% file: tab_benchmark_100.tex
\begin{table}[t]
\caption{\looseness - 1 
{\bf Benchmarking results ($d=100$ variables). }
Mean SID ($\downarrow$) and F1 score ($\uparrow$) with standard error of all methods on \num{30} random task instances. 
Methods in the top section use only observational data, in the bottom section both observational and interventional data.
We highlight the best result of each section and those within its \num{95}\% confidence interval according to an unequal variances $t$-test.
}
\label{tab:benchmark-100}
\vspace{10pt}
\centering
\begin{adjustbox}{max width=0.95\linewidth}
\begin{threeparttable}
\begin{tabular}{lcccccc}
\toprule
 &           \multicolumn{2}{c}{\domain{Linear}}            &            \multicolumn{2}{c}{\domain{Rff}}           &        \multicolumn{2}{c}{\domain{Grn}}       \\
\cmidrule{2-7}
Algorithm &              SID &               F1  &              SID &               F1 &              SID &               F1\\
\midrule
\ges               &  \highlight{2724.6 \err{362.2}} &  \highlight{0.471 \err{0.02}} &  \highlight{4703.3 \err{520.9}} &              0.240 \err{0.02} &  \highlight{6787.4 \err{351.8}} &  \highlight{0.031 \err{0.00}} \\ 
\lingam            &              6051.7 \err{585.1} &              0.150 \err{0.01} &              5489.9 \err{595.7} &              0.177 \err{0.02} &  \highlight{6726.5 \err{444.2}} &              0.011 \err{0.00} \\ 
\pc                &              5114.1 \err{621.1} &              0.287 \err{0.02} &              5294.8 \err{599.0} &              0.248 \err{0.03} &  \highlight{6894.4 \err{426.5}} &  \highlight{0.029 \err{0.00}} \\ 
\daggnn            &              6215.6 \err{598.9} &              0.101 \err{0.01} &              5445.5 \err{588.8} &              0.198 \err{0.02} &  \highlight{6574.3 \err{463.8}} &  \highlight{0.036 \err{0.01}} \\ 
\grandag           &              5307.5 \err{661.1} &              0.161 \err{0.02} &  \highlight{4522.0 \err{581.6}} &  \highlight{0.421 \err{0.04}}
& \highlight{6774.8 \err{419.3}} & \highlight{0.026 \err{0.01}}\\
\oursobserv &  \highlight{3213.8 \err{380.0}} &  \highlight{0.474 \err{0.04}} &  \highlight{3531.0 \err{498.0}} &  \highlight{0.506 \err{0.05}} &  \highlight{6661.2 \err{464.3}} &              0.000 \err{0.00} \\ 
\midrule
\gies              &  \highlight{1720.7 \err{306.1}} &  \highlight{0.639 \err{0.02}} &  \highlight{4528.3 \err{521.1}} &              0.257 \err{0.03} &              6691.2 \err{376.9} &              0.034 \err{0.00} \\ 
\igsp              &              4181.7 \err{478.3} &              0.316 \err{0.02} &              5544.7 \err{572.4} &              0.182 \err{0.02} &              6662.3 \err{401.4} &              0.027 \err{0.00} \\ 
\dcdi              &              5116.9 \err{525.4} &              0.105 \err{0.01} &  \highlight{3835.1 \err{413.3}} &              0.048 \err{0.00} &  \highlight{4410.8 \err{285.0}} &              0.027 \err{0.00} \\ 
\oursinterv &              2825.3 \err{379.0} &  \highlight{0.601 \err{0.03}} &  \highlight{3231.2 \err{500.2}} &  \highlight{0.550 \err{0.05}} &  \highlight{5237.5 \err{469.0}} &  \highlight{0.172 \err{0.05}} \\ 
\bottomrule
\end{tabular}
\end{threeparttable}
\end{adjustbox}
\vspace*{0pt}
\end{table}

%% file: tab_sachs.tex
\begin{table}[t]
\vspace*{5pt}
\caption{\looseness - 1 
{\bf Benchmarking results on the proteomics data by \citet{sachs2005causal}. }
We report the SHD ($\downarrow$), SID ($\downarrow$), and F1 score ($\uparrow$), and the number of edges predicted for all methods. 
Methods in the bottom section use the observational and interventional data, while the top row uses the concatenation of both, without the intervention targets.
We highlight the best result of each section.
}\label{tab:sachs}
\vspace{7pt}
\centering
\begin{adjustbox}{max width=\linewidth}
\begin{threeparttable}
\begin{tabular}{lcccc}
\toprule
 & SHD & SID & F1 & no.\ edges   \\
\midrule
\ges                 &   35 & \highlight{44} & 0.281 & 40 \\
\lingam              &   18 & 58 & 0.083 &  7 \\
\pc                 & 21 & 47 & 0.244 & 24 \\
\daggnn              &   26 & 49 & 0.273 & 27 \\
\grandag              &  \highlight{16}   & 38  & \highlight{0.473} & 21 \\
\textbf{\ours} (ours, trained on \lineardomain)\smash{${}^\|$} &   20 & 56 & 0.143 & 12 \\
\textbf{\ours} (ours, trained on \rffdomain)\smash{${}^\|$}   &   17 & 56 & 0.276 & 13 \\
\midrule
\gies                &   40 & \highlight{30} & 0.286 & 46 \\
\igsp                &   19 & 49 & 0.286 & 18 \\
\dcdi                &   \highlight{15} & 42 & \highlight{0.308} &  9 \\
\textbf{\ours} (ours, trained on \lineardomain)\smash{${}^\|$} &   20 & 50 & 0.250 & 11 \\
\textbf{\ours} (ours, trained on \rffdomain)\smash{${}^\|$}   &   16 & 49 & 0.267 & 15 \\
\bottomrule
\end{tabular}
\begin{tablenotes}
\footnotesize
\item[$\|$] Point estimate using decision threshold \num{0.5}
\end{tablenotes}
\end{threeparttable}
\end{adjustbox}
\vspace*{0pt}
\end{table}

%% file: fig_sachs_preds.tex
\newcommand{\figsachsheighttrue}{2.2cm} 
\newcommand{\figsachsheight}{1.55cm} 
\newcommand{\figsachsfillspace}{-2pt}

\begin{figure}[t]
    \centering
    \vspace*{10pt}
    \hspace*{-13pt}
    \begin{subfigure}[t]{0.19\textwidth}
        \centering
        \vspace*{-22pt}
        \includegraphics[height=\figsachsheighttrue]{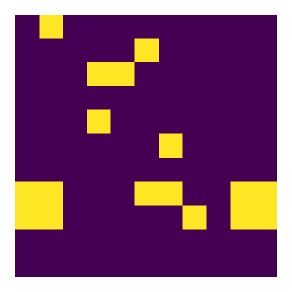}
        \vspace*{-3pt}
        \begin{center}
            $G^{\text{sachs}}$
        \end{center}
    \end{subfigure}
    \hspace*{-5pt}
    \begin{subfigure}[t]{0.80\textwidth}
        \centering
        \subfloat{
            \centering
            \includegraphics[height=\figsachsheight]{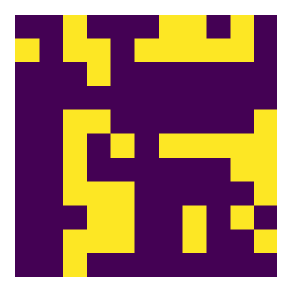}
            \hspace{\figsachsfillspace}
            \includegraphics[height=\figsachsheight]{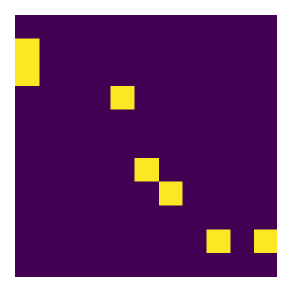}
            \hspace{\figsachsfillspace}
            \includegraphics[height=\figsachsheight]{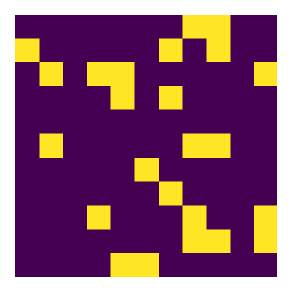}
            \hspace{\figsachsfillspace}            \includegraphics[height=\figsachsheight]{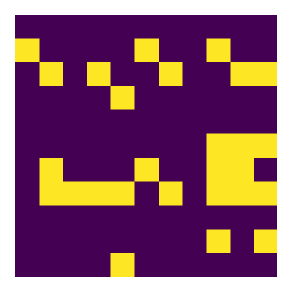}
            \hspace{\figsachsfillspace}            \includegraphics[height=\figsachsheight]{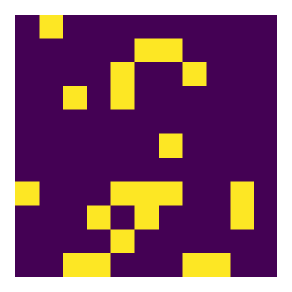}
            \hspace{\figsachsfillspace}        
            \includegraphics[height=\figsachsheight]{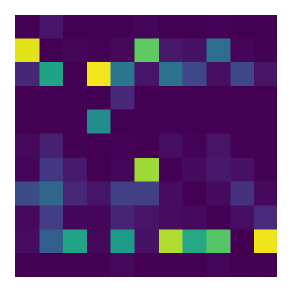}
            \hspace{\figsachsfillspace}       
            \includegraphics[height=\figsachsheight]{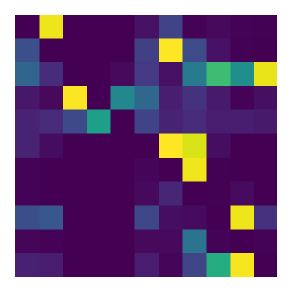}
        }
        \begin{center}\small
            \hspace{14pt}
            \textbf{\ges}\hspace*{18pt}
            \textbf{\lingam}\hspace*{20pt}
            \textbf{\pc}\hspace*{18pt}
            \textbf{\daggnn}\hspace*{5pt}
            \textbf{\grandag}\hspace*{7pt}
            $\substack{\text{\small\textbf{\ours}} \\ \text{(\lineardomain)}}$
            \hspace*{14pt}
            $\substack{\text{\small\textbf{\ours}} \\ \text{(\rffdomain)}}$
            \hspace*{14pt}
        \end{center}
        
        \hspace*{15pt}\subfloat{
            \centering
            \includegraphics[height=\figsachsheight]{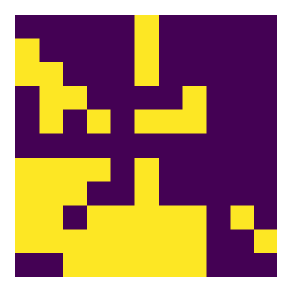}
            \hspace{\figsachsfillspace}
            \includegraphics[height=\figsachsheight]{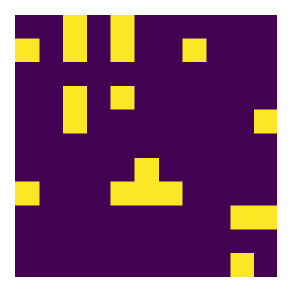}
            \hspace{\figsachsfillspace}
            \includegraphics[height=\figsachsheight]{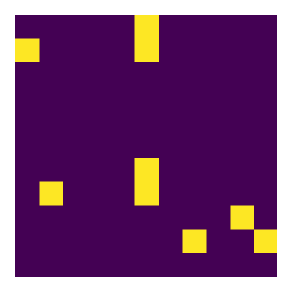}
            \hspace{\figsachsfillspace}            \includegraphics[height=\figsachsheight]{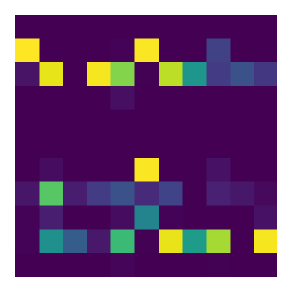}
            \hspace{\figsachsfillspace}            \includegraphics[height=\figsachsheight]{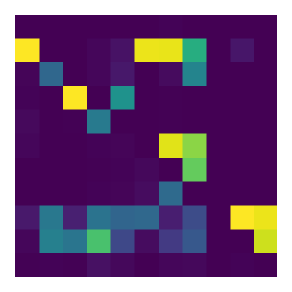}
        }
        \begin{center}\small
            \hspace{19pt}
            \textbf{\gies}\hspace{22pt}
            \textbf{\igsp}\hspace{24pt}
            \textbf{\dcdi}\hspace{20pt}
            $\substack{\text{\small\textbf{\ours}} \\ \text{(\lineardomain)}}$
            \hspace{17pt}
            $\substack{\text{\small\textbf{\ours}} \\ \text{(\rffdomain)}}$
            \hspace{20pt}
        \end{center}
    \end{subfigure}
    \vspace*{3pt}
    \caption{
    \looseness - 1
    {\bf Prediction of each method on the proteomics dataset by \citet{sachs2005causal}. }
    All baselines predict a point estimate of $G$, where edges are painted yellow.
    The posterior edge probabilities predicted by \ours, which were thresholded for Table \ref{tab:sachs}, are visualized as color gradients.
    The bottom row of methods use observational and interventional data, while the top row only uses the concatenation of both, without the intervention targets.
    The believed ground truth graph is shown on the left. 
    }
    \label{fig:sachs-predictions}
\end{figure}

%% file: fig_calibration_methods.tex
\newcommand{\figheightcalibmethod}{4.9cm} 
\newcommand{\figcalibmethodhspace}{-5pt} 
\newcommand{\figcalibmethodrelwidth}{0.32} 
\begin{figure}
    \centering
    \begin{subfigure}[t]{\figcalibmethodrelwidth\textwidth}
        \centering
        \subfloat{
            \centering
            \hspace*{\figcalibmethodhspace}\includegraphics[height=\figheightcalibmethod]{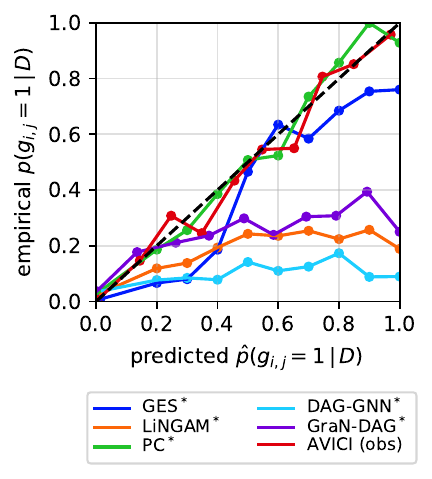}
        }
        
        \subfloat{
            \centering
            \hspace*{\figcalibmethodhspace}\includegraphics[height=\figheightcalibmethod]{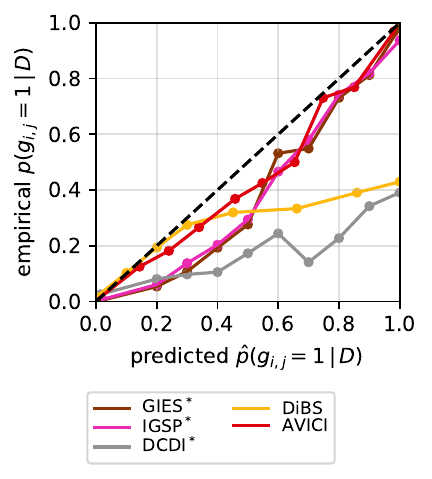}
        }
        \addtocounter{subfigure}{-2}
        \caption{\lineardomain}
    \end{subfigure}
    \hfill
    \begin{subfigure}[t]{\figcalibmethodrelwidth\textwidth}
        \centering
        \subfloat{
            \centering
            \hspace*{\figcalibmethodhspace}\includegraphics[height=\figheightcalibmethod]{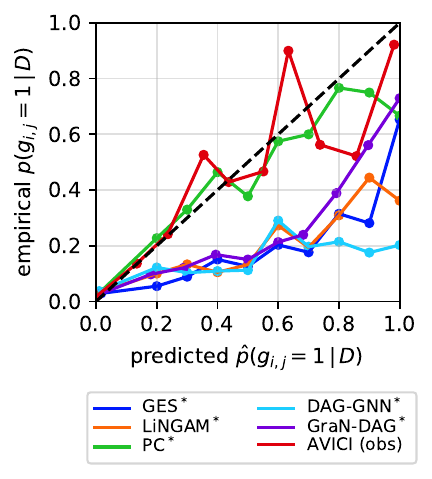}
        }
        
        \subfloat{
            \centering
            \hspace*{\figcalibmethodhspace}\includegraphics[height=\figheightcalibmethod]{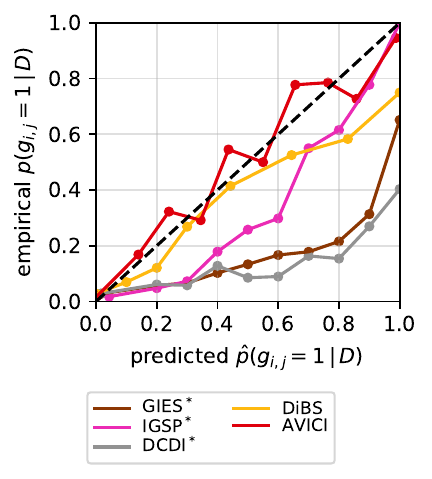}
        }
        \addtocounter{subfigure}{-2}
        \caption{\rffdomain}
    \end{subfigure}
    \hfill
    \begin{subfigure}[t]{\figcalibmethodrelwidth\textwidth}
        \centering
        \subfloat{
            \centering
            \hspace*{\figcalibmethodhspace}\includegraphics[height=\figheightcalibmethod]{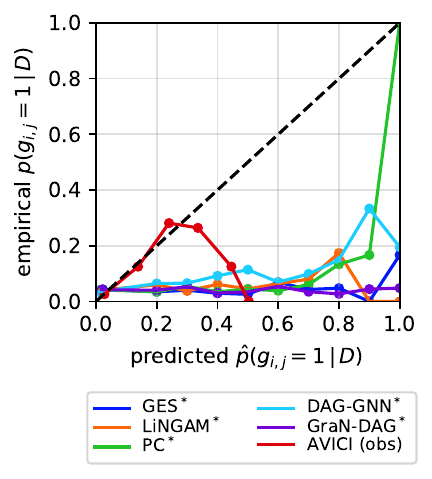}
        }
        
        \subfloat{
            \centering
            \hspace*{\figcalibmethodhspace}\includegraphics[height=\figheightcalibmethod]{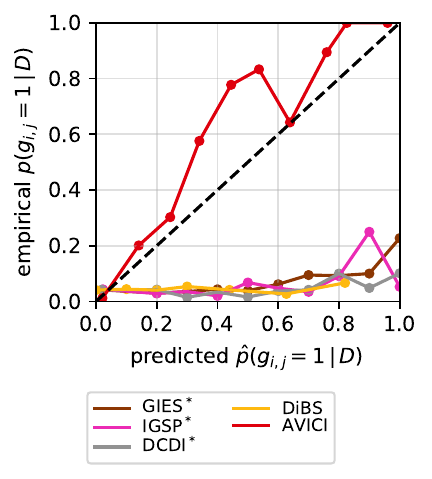}
        }
        \addtocounter{subfigure}{-2}
        \caption{\grndomain}
    \end{subfigure}
    \caption{
    \looseness - 1
    {\bf Calibration plots ($d=30$) }
    for all methods in the experiments of Section \ref{ssec:benchmarking} and Table \ref{fig:calibration-avici}b.
    The top and bottom rows of plots show the methods that use observational data and a mix of observational and interventional data, respectively.
    Methods with an asterix use the nonparametric DAG bootstrap to estimate edge probabilities \citep{friedman1999data} (cf.\ Section \ref{ssec:benchmarking}).
    \vfigurecaptionbelow
    }
    \label{fig:calibration-all}
\end{figure}

%% file: tab_benchmark_30_prob.tex
\begin{table}[t]
\caption{\looseness - 1 
{\bf Probabilistic metrics for the benchmark ($d=30$ variables). }
Mean AUROC ($\uparrow$) and AUPRC ($\uparrow$) with standard error of all methods on ten random task instances. 
Methods in the top section use only observational data, in the bottom section both observational and interventional data.
We highlight the best result of each section and those within its \num{95}\% confidence interval according to an unequal variances $t$-test.
}\label{tab:benchmark-30-prob}
\vspace{7pt}
\centering
\begin{adjustbox}{max width=0.95\linewidth}
\begin{threeparttable}
\begin{tabular}{lcccccc}
\toprule
 &           \multicolumn{2}{c}{\domain{Linear}}            &            \multicolumn{2}{c}{\domain{Rff}}           &        \multicolumn{2}{c}{\domain{Grn}}       \\
\cmidrule{2-7}
Algorithm &              AUROC &               AUPRC  &              AUROC &               AUPRC &              AUROC &               AUPRC\\
\midrule
\ges${}^*$       &              0.930 \err{0.01} &  \highlight{0.643 \err{0.05}} &  \highlight{0.759 \err{0.04}} &              0.289 \err{0.06} &  \highlight{0.496 \err{0.02}} &              0.045 \err{0.00} \\
\lingam${}^*$    &              0.752 \err{0.06} &              0.365 \err{0.10} &              0.701 \err{0.04} &              0.229 \err{0.03} &  \highlight{0.537 \err{0.03}} &              0.057 \err{0.01} \\
\pc${}^*$        &              0.771 \err{0.04} &              0.469 \err{0.06} &  \highlight{0.825 \err{0.04}} &  \highlight{0.507 \err{0.07}} &  \highlight{0.510 \err{0.02}} &              0.052 \err{0.01} \\
\daggnn${}^*$    &              0.621 \err{0.03} &              0.097 \err{0.02} &              0.693 \err{0.03} &              0.174 \err{0.01} &  \highlight{0.547 \err{0.05}} &              0.082 \err{0.03} \\
\grandag${}^*$ &                0.685 \err{0.04} &              0.222 \err{0.03} &  \highlight{0.781 \err{0.04}} &  \highlight{0.419 \err{0.09}} & \highlight{0.534 \err{0.08}} & \highlight{0.113 \err{0.05}} \\
\oursobserv      &  \highlight{0.979 \err{0.01}} &  \highlight{0.767 \err{0.06}} &  \highlight{0.801 \err{0.07}} &  \highlight{0.571 \err{0.12}} &  \highlight{0.678 \err{0.10}} &  \highlight{0.185 \err{0.04}} \\
\midrule
\gies${}^*$      &  \highlight{0.981 \err{0.01}} &  \highlight{0.879 \err{0.04}} &  \highlight{0.769 \err{0.06}} &              0.389 \err{0.08} &              0.517 \err{0.03} &              0.070 \err{0.01} \\
\igsp${}^*$      &              0.942 \err{0.01} &              0.660 \err{0.05} &  \highlight{0.822 \err{0.04}} &              0.374 \err{0.07} &              0.471 \err{0.02} &              0.049 \err{0.01} \\
\dcdi${}^*$      &              0.771 \err{0.03} &              0.306 \err{0.03} &  \highlight{0.740 \err{0.05}} &              0.283 \err{0.06} &              0.557 \err{0.07} &              0.113 \err{0.05} \\
\dibs             &              0.837 \err{0.03} &              0.524 \err{0.05} &  \highlight{0.740 \err{0.05}}      &              0.340 \err{0.05} &              0.517 \err{0.03} &              0.048 \err{0.01} \\
\oursinterv      &  \highlight{0.987 \err{0.01}} &  \highlight{0.902 \err{0.04}} &  \highlight{0.862 \err{0.05}} &  \highlight{0.669 \err{0.10}} &  \highlight{0.901 \err{0.04}} &  \highlight{0.656 \err{0.10}} \\
\bottomrule
\end{tabular}
 \begin{tablenotes}
\footnotesize
\item[$*$] Nonparametric DAG bootstrap \citep{friedman1999data}
 \end{tablenotes}
\end{threeparttable}
\end{adjustbox}
\vfigurecaptionbelow
\vspace*{0pt}
 \end{table}